\documentclass[12pt]{article}
\usepackage[utf8]{inputenc}
\usepackage[T1]{fontenc}
\usepackage{amssymb}
\usepackage{amsmath}
\usepackage{amsfonts}
\usepackage{stfloats}
\usepackage [autostyle, english = american]{csquotes}
\MakeOuterQuote{"}
\usepackage{subcaption}
\usepackage{textcomp}
\usepackage{blindtext}
\usepackage{graphicx}
\usepackage{lettrine}
\usepackage{lipsum}
\usepackage{float}
\usepackage{multicol}
\usepackage{multirow}
\usepackage{textcomp}
\usepackage{layouts}
\usepackage[font=small,skip=0pt]{caption}
\usepackage{color}
\usepackage{mathtools}
\usepackage{booktabs}

\usepackage{array}
\newcolumntype{L}[1]{>{\raggedright\let\newline\\\arraybackslash\hspace{0pt}}m{#1}}
\newcolumntype{C}[1]{>{\centering\let\newline\\\arraybackslash\hspace{0pt}}m{#1}}
\newcolumntype{R}[1]{>{\raggedleft\let\newline\\\arraybackslash\hspace{0pt}}m{#1}}
\usepackage[linesnumbered]{algorithm2e}
\usepackage{bbm}
\usepackage{duckuments}
\usepackage[hidelinks]{hyperref}
\usepackage[left=2cm, right=3cm, top=2cm, bottom=2.5cm]{geometry}
\usepackage{xcolor}
\definecolor{myred}{rgb}{0.7529,0,0}  
\usepackage{authblk}
\usepackage[numbers]{natbib}

\title{High-Gain Disturbance Observer for Robust Trajectory Tracking of Quadrotors}

\author[1]{Mohammadreza Izadi}
\author[1]{Reza Faieghi\thanks{Corresponding author: reza.faieghi@torontomu.ca}}
\affil[1]{Department of Aerospace Engineering, Toronto Metropolitan University, Toronto, Canada}

\date{}

\begin{document}
\maketitle

\begin{abstract}
This paper presents a simple method to boost the robustness of quadrotors in trajectory tracking.
The presented method features a high-gain disturbance observer (HGDO) that provides disturbance estimates in real-time.
The estimates are then used in a trajectory control law to compensate for disturbance effects.
We present theoretical convergence results showing that the proposed HGDO can quickly converge to an adjustable neighborhood of actual disturbance values.
We will then integrate the disturbance estimates with a typical robust trajectory controller, namely sliding mode control (SMC), and present Lyapunov stability analysis to establish the boundedness of trajectory tracking errors.
However, our stability analysis can be easily extended to other Lyapunov-based controllers to develop different HGDO-based controllers with formal stability guarantees.
We evaluate the proposed HGDO-based control method using both simulation and laboratory experiments in various scenarios and in the presence of external disturbances.
Our results indicate that the addition of HGDO to a quadrotor trajectory controller can significantly improve the accuracy and precision of trajectory tracking in the presence of external disturbances.
\end{abstract}


\section{Introduction}\label{se:intro}
Recently, quadrotor uncrewed aerial vehicles (UAVs) have garnered significant attention from researchers due to their potential applications, including tasks like power line monitoring, inspection, logistics distribution, and firefighting. To execute these complex missions accurately and with high quality, it is crucial to ensure the stability and robustness of the position and attitude control systems. However, achieving robustness is a challenging task when dealing with external disturbances. Quadrotors are susceptible to various external disturbances, including wind gusts \cite{galway2008modeling}, airflow distortion in the vicinity of surfaces \cite{he2020quasi}, and wake turbulence \cite{nathanael2022numerical}. These disturbances are often unmodeled or challenging to measure. Consequently, enhancing reliability and safety in trajectory-tracking missions has emerged as a formidable challenge.

Various control approaches are studied for robust trajectory tracking of vehicles in the presence of disturbances. Examples include model predictive control \cite{greeff2018flatness, zhang2023finite, nan2022nonlinear}, sliding mode control (SMC) \cite{xu2006sliding, hou2020nonsingular, zheng2014second, kabiri20193d}, adaptive control \cite{ de2013position, song20231}, neural networks \cite{donti2020enforcing}, and reinforcement learning \cite{han2022cascade}.


Numerous control methods exist that rely on disturbance observers (DOs). 
For example, the time delay controller (TDC) involves a DO that uses the time delay between the control input and the system output to estimate the disturbance \cite{youcef1990time}, successfully applied to quadrotor attitude control, altitude control, and position control \cite{lim2014altitude}. 
Sliding mode DO is another variation of DOs that relies on a sliding mode observer to estimate the disturbance, improving the vehicle robustness to external disturbances and sensor noise \cite{hall2006sliding}. 
Further examples include generalized extended state observer \cite{shi2018generalized}, and the uncertainty and disturbance estimator (UDE) \cite{talole2008model, chandar2014improving}.
The latter offers several advantages, including the absence of system delays, the elimination of control signal oscillations, and the obviation of the need to measure state vector derivatives \cite{zhong2004control}.


The challenge of disturbances in constrained systems has been addressed by using iterative learning control, taking into account both input and output constraints, along with model uncertainty and output disturbances \cite{zhou2022robust}.
The use of DO in a hierarchical control framework combined with adaptive control techniques is investigated in  \cite{liang2021geometric}, enabling quadrotors to adapt to varying disturbances and compensate for aerodynamic damping effects, resulting in robust and precise control. 
A bank of nonlinear DOs is utilized alongside a set of generalized backstepping and SMCs to counteract the impact of unaccounted uncertainties that affect the vehicle during flight \cite{fethalla2018robust}.
Finite-time nonlinear DO is studied in \cite{huang2022finite}.
An integrated adaptive dynamic programming (ADP) technique is used in \cite{stojanovic2023fault} to achieve asymptotic tracking. By using real-time input-output data, the control algorithm can compute an approximated optimal fault-tolerant control. This approach allows the system to reject disturbances and maintain stable performance even in the presence of uncertainties and faults.

While DO-based control is proven to be effective for quadrotor trajectory tracking, many existing DOs suffer from complex structures that add to the computational overhead of flight controllers. While significant progress has been made on the computational power of flight controllers, the extensive computational demands for autonomous or semi-autonomous operations, coupled with the power and weight restrictions of quadrotors, impose constraints on the available computing power of flight controllers; therefore, simple and computationally efficient DOs are still desired.
Also, complex DOs often involve several tuning parameters which require an involved tuning process to ensure a fast convergence rate.

One category of nonlinear observers that are simple and fast is high-gain observers (HGOs) \cite{khalil2017high}.
As its name suggests, HGO relies on the idea of applying a high gain to quickly recover the state estimates.
HGOs present several desirable properties.
First, they are relatively simple to design and implement since the observer is a copy of the model of the system with a gain whose expression is explicitly given.
Second, the observer tuning is realized simply through the choice of a single scalar design parameter.
Finally, they can provide global or semi-global stability results for a large class of systems.
Such appealing properties; however, come at a cost. 
Conventional HGOs suffer from measurement noise amplification; however, this problem is alleviated by recent designs \cite{boizot2010adaptive, won2015high}.

Given the advantages of HGOs, their applications for disturbance estimation are also explored, giving rise to high-gain disturbance observers (HGDOs) \cite{khalil2017extended}. 
However, it has not been thoroughly studied for quadrotors, especially on an actual vehicle with real measurement errors and real disturbances, beyond simulation. 
The papers that we were able to find on this topic either presented a limited simulation study or focused on only attitude control \cite{boss2017uncertainty, zhao2019composite, lu2020uncertainty, shi2018generalized}. Another related work is \cite{boss2021high} which does not design HGDO, but rather an HGO for quadrotor state estimation, as an alternative to the extended Kalman filter.
It is worth noting that the application of HGDOs for helicopters has been studied in a few research papers \cite{lee2021output, fang2016robust}. While helicopters and quadrotors are both rotary-wing aircraft, there exist fundamental differences in their flight mechanisms and control. Moreover, the tail rotor in helicopters adds an additional degree of authority for lateral control, which is missing in quadrotors.

In light of the above discussion, our objective here is to design HGDO for trajectory control of quadrotors.
The main contribution of this paper is that it develops HGDO for the attitude and position control of quadrotors. We integrated the proposed HGDO with a Lypaunov-based robust control law, namely SMC. We conduct extensive simulation and hardware experiments to compare the proposed HGDO+SMC method with existing methods. Our results demonstrate fast and accurate disturbance estimation, enabling accurate trajectory tracking for quadrotors, outperforming the benchmark methods tested.  
We present Lyapunov stability analysis to establish the boundedness of tracking errors and disturbance estimations.
The SMC can be easily replaced by another Lyapunov-based controller, and a similar stability analysis can be derived for the alternative controller. 
Overall, our intention is not a major overhaul in the quadrotor flight control, but rather to introduce a simple, easy-to-tune, and computationally efficient module that can be added to a flight controller and boost trajectory tracking robustness against external disturbances.

\section{Quadrotor Modeling}\label{sec:modeling}
The details of the quadrotor model are explained in various references such as \cite{mahony2012multirotor}. 
Here, we attempt to write the quadrotor translation and rotational dynamics in second-order controllability canonical forms which will become useful in our developments.
The problem statement will follow.

\begin{figure}[t]
    \centering
    \includegraphics[width = 0.6\linewidth]{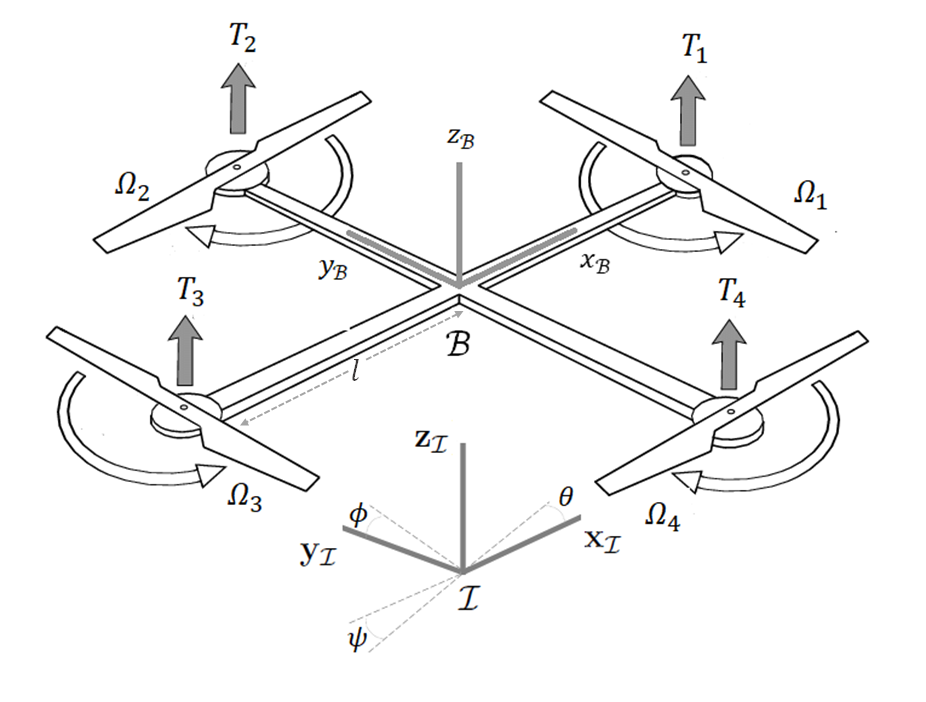}
    \caption{Quadrotor model and coordinate frames}
    \label{fig:modeling}
\end{figure}

Let us begin by setting $\mathcal{I}=\left\{{\bf{x}}_\mathcal{I},{\bf{y}}_\mathcal{I},{\bf{z}}_\mathcal{I} \right\}$ as an Earth-fixed inertial coordinates frame, and $\mathcal{B}=\left\{{\bf{x}}_\mathcal{B},{\bf{y}}_\mathcal{B},{\bf{z}}_\mathcal{B} \right\}$ as the body-fixed coordinates frame whose origin coincides with the center of mass of the vehicle (Fig. \ref{fig:modeling}). 
We assume that the quadrotor body is rigid and symmetric, with arms aligned to ${\bf{x}}_\mathcal{B}$ and ${\bf{y}}_\mathcal{B}$.
The length of each arm is $l$, and the mass of the vehicle is $m$.
Also, the inertia matrix is ${\bf{J}}$ which is diagonal ${\bf{J}} = {\rm{diag}}\left(J_x,J_y,J_z\right)$ due to the symmetry of the vehicle.

We denote the position of the vehicle in $\mathcal{I}$ by ${\boldsymbol{\xi}}=\left[x,y,z\right]^T$.
For the vehicle attitude, we use ${\boldsymbol{\eta}}=\left[\phi, \theta,\psi\right]^T$, where $ - \pi  < \phi  \le \pi $, $ - \frac{\pi }{2} \le \theta  \le \frac{\pi }{2}$, and $ - \pi  < \psi  \le \pi $ are the Euler angles representing pitch, roll, and yaw in the yaw-pitch-roll sequence.
With the above Euler angle configuration, the rotation matrix from $\mathcal{B}$ to $\mathcal{I}$ takes the following form
\begin{equation}
\label{3}
{\bf{R}} = \left[ {\begin{array}{*{20}{c}}
{c\theta c\psi }&{s\phi s\theta c\psi  - c\phi s\psi }&{c\phi s\theta c\psi  + s\phi s\psi }\\
{c\theta s\psi }&{s\phi s\theta s\psi  + c\phi c\psi }&{c\phi s\theta s\psi  - s\phi c\psi }\\
{ - s\theta }&{s\phi c\theta }&{c\phi c\theta }
\end{array}} \right],
\end{equation}
where $c$ and $s$ stand for cosine and sine functions.
Also, if ${\boldsymbol{\omega}}=\left[p,q,r\right]^T$ represents the angular velocity vector, then according to the Euler kinematical equation, we have $\dot {\boldsymbol{\eta}} = {\bf{H}}\left({\boldsymbol{\eta}}\right) {\boldsymbol{\omega}}^\mathcal{B}$, where the superscript $\mathcal{B}$ indicates that the vector components are expressed in $\mathcal{B}$ and 
\begin{equation}\label{eq:H}
{\bf{H}}\left( {\boldsymbol{\eta }} \right) = \left[ {\begin{array}{*{20}{c}}
1&{\sin \phi \tan \theta }&{\cos \phi \tan \theta }\\
0&{\cos \phi }&{ - \sin \phi }\\
0&{{{\sin \phi } \mathord{\left/
 {\vphantom {{\sin \phi } {\cos \theta }}} \right.
 \kern-\nulldelimiterspace} {\cos \theta }}}&{{{\cos \phi } \mathord{\left/
 {\vphantom {{\cos \phi } {\cos \theta }}} \right.
 \kern-\nulldelimiterspace} {\cos \theta }}}
\end{array}} \right].
\end{equation}

Each of the vehicle's rotors produces a thrust $T_i,\;i=\left\{1,2,3,4\right\}$ in the direction of $\bf{z}_\mathcal{B}$. 
$T_i$s are usually approximated by $k_T \Omega_i^2$ where $\Omega_i$ is the angular velocity of $i$-th rotor, and $k_T$ is a coefficient.
The rotor angular velocities on the $\bf{x}_\mathcal{B}$ and $\bf{y}_\mathcal{B}$ axes have opposite signs ($\Omega_{1,3} > 0$, $\Omega_{2,4} < 0$) to counterbalance the reaction torque induced by the rotors and to control $\psi$.
Let ${\bf{f}}_A$ and ${\boldsymbol{\tau}}_A$ be the aerodynamic force and torque vectors produced by $T_i$s.
Then, we can express them in $\mathcal{B} $ as follows
\begin{equation}
{\bf{f}}_A^{\cal B} = \left[ {\begin{array}{*{20}{c}}
0\\
0\\
T
\end{array}} \right],\;{\rm{and}}\;{\boldsymbol{\tau}}_A^\mathcal{B} = \left[ {\begin{array}{*{20}{c}}
{lk_T\left( {\Omega _4^2 - \Omega _2^2} \right)}\\
{lk_T\left( {\Omega _3^2 - \Omega _1^2} \right)}\\
{lk_Q\left( {\Omega _1^2 - \Omega _2^2 + \Omega _3^2 - \Omega _4^2} \right)}
\end{array}} \right],
\end{equation}
where $l$ is the distance from the center of mass to the rotor, $T={k_T}\sum\nolimits_{i = 1}^4 {\Omega _i^2}$ is the total thrust, $k_T$ is thrust coefficient and $k_Q$ is torque coefficient. The efficiency of the propulsion system is greatly affected by the thrust and torque coefficients, which are shaped by the rotor's design characteristics. These coefficients are also affected by the motion of the vehicle as detailed in  \cite{bouabdallah2007full, bouabdallah2006towards}.

In addition, we assume that there exist two unknown disturbance vectors ${\bf{d}}_{{\boldsymbol{\xi}}}=\left[d_x,d_y,d_z\right]^T$ and ${\bf{d}}_{{\boldsymbol{\eta}}}=\left[d_\phi,d_\theta,d_\psi\right]^T$.

Applying Newton's law of motion, the translational dynamics of the vehicle take the following form
\begin{equation}\label{eq:transDyn}
    {\ddot {\boldsymbol{\xi}}  =  - {{\bf{g}}} + \frac{1}{m}\left( {{\bf{f}}_A^{\cal I} + {\bf{d}}_{\boldsymbol{\xi }}^{\cal I}} \right),}
\end{equation}
where ${{\bf{f}}_A^{\cal I} = {\bf{Rf}}_A^{\cal B}}$, ${{\bf{d}}_{\boldsymbol{\xi }}^{\cal I} = {\bf{Rd}}_{\boldsymbol{\xi }}^{\cal B}}$ and ${{\bf{g}}} = \left[0\;0\;g\right]^T$ is the gravity vector with $g$ set to $9.81 m/s^2$.
Using Euler's rotation theorem, the rotational dynamics of the vehicle take the following form
\begin{equation}\label{eq:rotDyn}
    \dot{\boldsymbol{\omega}}^{\cal B} = {\bf{J}}^{ - 1}\left( -{\boldsymbol{\omega}} ^{\cal B} \times {\bf{J}}{\boldsymbol{\omega}} ^{\cal B} + {\boldsymbol{\tau}}_A^{\cal B} + {\bf{d}}_{\boldsymbol{\xi }}^{\cal B} \right),\;{\rm{and}}\;\dot {\boldsymbol{\eta}}  = {\bf{H}}\left( {{\boldsymbol{\eta}} } \right){\boldsymbol{\omega}} ^{\cal B},
\end{equation}
where the index $\times$ indicates the cross product.

We can now use \eqref{eq:transDyn} and \eqref{eq:rotDyn} to write the quadrotor dynamics as a set of second-order nonlinear systems in controllability canonical forms.
However, due to the particular structure of \eqref{eq:rotDyn}, the equations will become complicated, and this will lead to complex DO and control laws.
If $\boldsymbol{\phi}$ and $\boldsymbol{\theta}$ are small, $\bf{H}$ can be greatly simplified such that $\dot{\boldsymbol{\eta}} \approx {\boldsymbol{\omega}}$.
As such, the expressions in \eqref{eq:transDyn} and \eqref{eq:rotDyn} transfer into

\begin{equation}
\label{eq:states}
\left\{ {\begin{array}{*{20}{l}}
{\ddot x = (\cos \varphi \sin \theta \cos \psi  + \sin \varphi \sin \psi )\frac{1}{m}{u_1} + {d_x}},\\
{\ddot y = (\cos \varphi \sin \theta \sin \psi  - \sin \varphi \sin \psi )\frac{1}{m}{u_1}+ {d_y}},\\
{\ddot z =  - g + (\cos \varphi \cos \theta )\frac{1}{m}{u_1}+ {d_z}},\\
{\ddot \phi  = \dot \theta \dot \psi (\frac{{{I_y} - {I_z}}}{{{I_x}}})   + \frac{1}{{{I_x}}}{u_2} + {d_\phi }},\\
{\ddot \theta  = \dot \varphi \dot \psi (\frac{{{I_z} - {I_x}}}{{{I_y}}})   + \frac{1}{{{I_y}}}{u_3} + {d_\theta }},\\
{\ddot \psi  = \dot \varphi \dot \theta (\frac{{{I_x} - {I_y}}}{{{I_z}}}) + \frac{1}{{{I_z}}}{u_4} + {d_\psi }}.\\
\end{array}} \right.
\end{equation}

By setting
 ${{\bf{x}}_1} = {\boldsymbol{\xi}}$, ${{\bf{x}}_2} = \dot{\boldsymbol{\xi}}$, ${{\bf{x}}_3} = {\boldsymbol{\eta}}$, ${{\bf{x}}_4} = \dot{\boldsymbol{\eta}}$,  ${\bf{d_1}}=\frac{{\bf{d_{\boldsymbol{\xi}}}}}{m}$, ${\bf{d_2}}={\bf{J}}^{-1}{\bf{d_{\boldsymbol{\eta}}}}$, ${{\bf{u}}_1} = {\frac{1}{m}}{\bf{b}}{{u}}_1$, ${\bf{u}}_2=[u_2/I_x, u_3/I_y, u_4/I_z]^T$, ${\bf{g}} = {[0,0,g]^T}$, 
\begin{equation*}
{\bf{b}} = \left[ {\begin{array}{*{20}{c}}
{c\varphi s\theta c\psi  + s\varphi s\psi }\\
{c\varphi s\theta s\psi  - s\varphi s\psi }\\
{c\varphi c\theta }
\end{array}} \right],\;{\rm{and}}\;{{\bf{f}}_2}({{\bf{x}}_4}) = \left[ {\begin{array}{*{20}{c}}
{\frac{{{J_y} - {J_z}}}{{{J_x}}}\dot \theta \dot \psi }\\
{\frac{{{J_z} - {J_x}}}{{{J_y}}}\dot \phi \dot \psi }\\
{\frac{{{J_x} - {J_y}}}{{{J_z}}}\dot \phi \dot \theta }
\end{array}} \right],
\end{equation*}
the equations in \eqref{eq:states} can be converted into the following state space representation
\begin{equation}
\label{eq:statespaceform}
\left\{ {\begin{array}{*{20}{l}}
{{{{\bf{\dot x}}}_1} = {{\bf{x}}_2}},\\
{{{{\bf{\dot x}}}_2} = - {\bf{g}} + {{\bf{u}}_1} + {{\bf{d}}_1}},\\
{{{{\bf{\dot x}}}_3} = {{\bf{x}}_4}},\\
{{{{\bf{\dot x}}}_4} = {{\bf{f}}_2}({{\bf{x}}_4}) + {{\bf{u}}_2} + {{\bf{d}}_2}},
\end{array}} \right.
\end{equation}



Our first objective is to design an HGDO that can estimate ${\bf{d}}_1$ and ${\bf{d}}_2$.
Our second objective is to use the disturbance estimates and design control laws ${\bf{u}}_1$ and ${\bf{u}}_2$ such that the vehicle can follow a desired trajectory $\left[{\bf{x}}_{1d}^T, {\bf{x}}_{3d}^T\right]^T$ in the presence of unknown disturbances. 
Once ${\bf{u}}_1$ and ${\bf{u}}_2$ are determined, $\Omega_i$ can be calculated using the following expression

\begin{equation}\label{eq:motorMixing}
\left[ {\begin{array}{*{20}{c}}
{\Omega _1^2}\\
{\Omega _2^2}\\
{\Omega _3^2}\\
{\Omega _4^2}
\end{array}} \right] = \left[ {\begin{array}{*{20}{c}}
{\frac{1}{{4{k_T}}}}&0&{ - \frac{1}{{2l{k_T}}}}&{ - \frac{1}{{4{k_Q}}}}\\
{\frac{1}{{4{k_T}}}}&{\frac{1}{{2l{k_T}}}}&0&{\frac{1}{{4{k_Q}}}}\\
{\frac{1}{{4{k_T}}}}&0&{\frac{1}{{2l{k_T}}}}&{ - \frac{1}{{4{k_Q}}}}\\
{\frac{1}{{4{k_T}}}}&{ - \frac{1}{{2l{k_T}}}}&0&{\frac{1}{{4{k_Q}}}}
\end{array}} \right]\left[ {\begin{array}{*{20}{c}}
u_1\\
u_2\\
u_3\\
u_4

\end{array}} \right].
\end{equation}

We assume that the disturbance terms ${\bf{d}}_1$ and ${\bf{d}}_2$ and their derivatives are bounded. Let us denote the $j$-th component of ${\bf{d}}_i,\;i=\left\{1,2\right\}$ by $d_i^j$. Then,
\begin{equation}\label{eq:assumption}
\left\| \dot{d}_i^j \right\| \le \delta_i^j   ,
\end{equation}
where $\delta_i^j$s denote unknown but finite positive constants and $\left\| \cdot \right\|$ is the $\mathcal{L}_1$ norm defined as ${\left\| {\chi\left( t \right)} \right\|} = \int_0^t {|\chi\left( \alpha  \right)|d\alpha } $.
The disturbance terms can generally include unknown external disturbances, gyroscopic effects of rotors, or aerodynamic effects such as drag. 
The models for gyroscopic effects or drag exist in the literature \cite{martini2022euler} and one can include them in \eqref{eq:transDyn} and \eqref{eq:rotDyn} to have a more elaborated model.
However, these effects can be combined into disturbance terms in each axes, and HGDO can estimate their amplitudes.
Therefore, one advantage of using HGDO is reducing the need for complex models.


\section{High-Gain Disturbance Observer Design}\label{se:HGDO}
Our objective in this section is to design an HGDO to estimate ${\bf{d}}_1$ and ${\bf{d}}_2$.
Let ${\hat{\bf{d}}_1}$ and ${\hat{\bf{d}}_2}$ be the estimated values of ${\bf{d}}_1$ and ${\bf{d}}_2$.
The basic idea behind HGDO is to construct an observer of the following form
\begin{equation}\label{eq:HGOIdaea}
\left\{ {\begin{array}{*{20}{c}}
{{{{\bf{\dot{\hat d}}}}_1} = \frac{1}{{{\varepsilon _1}}}\left( {{{\bf{d}}_1} - {{{\bf{\hat d}}}_1}} \right)},\\
{{{{\bf{\dot{\hat d}}}}_2} = \frac{1}{{{\varepsilon _2}}}\left( {{{\bf{d}}_2} - {{{\bf{\hat d}}}_2}} \right),}
\end{array}} \right.
\end{equation}
where $\frac{1}{\varepsilon_1}$ and $\frac{1}{\varepsilon_2}$ are the observer gains.
Each equation in \eqref{eq:HGOIdaea} constitutes a first-order filter of the form $\frac{1}{{{\varepsilon _i}s + 1}}$.
By choosing small positive values for ${\varepsilon_i}$, the settling time of the filter becomes small, and therefore $\hat{\bf{d}}_i$ quickly converges to ${\bf{d}}_i$. 

Note that ${\bf{d}}_i$s are unknown; however, from \eqref{eq:statespaceform}, they can be expressed as follows 
\begin{equation}\label{eq:d}
\left\{ {\begin{array}{*{20}{l}}
{{{\bf{d}}_1} = {{{\bf{\dot x}}}_2} + {{\bf{g}}} - {{\bf{u}}_1}},\\
{{{\bf{d}}_2} = {{{\bf{\dot x}}}_4} - {{\bf{f}}_2}({{\bf{x}}_4}) - {{\bf{u}}_2}}.
\end{array}} \right.
\end{equation}
Therefore, one can suggest the following HGDO structure for quadrotor
\begin{equation}\label{eq:dhat}
\left\{ {\begin{array}{*{20}{l}}
{{{{\bf{\dot{\hat d}}}}_1} = \frac{1}{{{\varepsilon _1}}}\left( {{{{\bf{\dot x}}}_2} + {{\bf{g}}} - {{\bf{u}}_1} - {{{\bf{\hat d}}}_1}} \right)},\\
{{{{\bf{\dot{\hat d}}}}_2} = \frac{1}{{{\varepsilon _2}}}\left( {{{{\bf{\dot x}}}_4} - {{\bf{f}}_2}({{\bf{x}}_4}) - {{\bf{u}}_2} - {{{\bf{\hat d}}}_2}} \right).}
\end{array}} \right.
\end{equation}

We assume that the initial disturbance estimates are set to zero i.e. ${{\bf{\hat d}}_i}\left( 0 \right) = 0$.
The drawback of \eqref{eq:dhat} is the inclusion of derivatives of system states which amplifies the effect of measurement noise.
Inspired by \cite{won2015high}, we propose an HGDO using auxiliary varibles
\begin{equation}\label{eq:gamma}
{{\boldsymbol{\gamma }}_1} = {\hat {\bf{d}}_1} - \frac{{{{\bf{x}}_{\bf{2}}}}}{{{\varepsilon _1}}},\;{\rm{and}}\;{{\boldsymbol{\gamma }}_2} = {\hat {\bf{d}}_2} - \frac{{{{\bf{x}}_{\bf{4}}}}}{{{\varepsilon _2}}},
\end{equation}
with dynamics given as follows
\begin{equation}\label{eq:gammaDot}
\left\{ {\begin{array}{*{20}{c}}
{{{{\boldsymbol{\dot \gamma }}}_1} =  - \frac{1}{{{\varepsilon _1}}}\left( {{{\boldsymbol{\gamma }}_1} + \frac{{{{\bf{x}}_2}}}{{{\varepsilon _1}}}} \right) + \frac{1}{{{\varepsilon _1}}}\left( {{\bf{g}}^{\cal I} - {{\bf{u}}_1}} \right)},\\
{{{{\boldsymbol{\dot \gamma }}}_2} =  - \frac{1}{{{\varepsilon _2}}}\left( {{{\boldsymbol{\gamma }}_2} + \frac{{{{\bf{x}}_4}}}{{{\varepsilon _2}}}} \right) + \frac{1}{{{\varepsilon _2}}}\left( { - {{\bf{f}}_2}({{\bf{x}}_4}) - {{\bf{u}}_2}} \right).}
\end{array}} \right.
\end{equation}

To establish the convergence results for the observer, let us define the disturbance estimation error as
${\bf{\tilde d}}_i = {\bf{d}}_i - {\bf{\hat d}}_i$.
Taking the derivative of ${\bf{\tilde d}}_i$ and using \eqref{eq:gamma} result in
\begin{equation}\label{eq:estimationErrDyn}
{{\bf{\dot{\tilde d}}}_1} = {{\bf{\dot d}}_1} - \left( {{{{\boldsymbol{\dot \gamma }}}_1} + \frac{{{{{\bf{\dot x}}}_2}}}{{{\varepsilon _1}}}} \right),\;{{\bf{\dot {\tilde d}}}_2} = {{\bf{\dot d}}_2} - \left( {{{{\boldsymbol{\dot {\gamma }}}}_2} + \frac{{{{{\bf{\dot x}}}_4}}}{{{\varepsilon _2}}}} \right).
\end{equation}
Substituting \eqref{eq:gammaDot} in \eqref{eq:estimationErrDyn} leads to
\begin{equation}\label{eq:estimationErrDyn2}
    {{\bf{\dot{\tilde d}}}_i} =  - \frac{1}{{{\varepsilon _i}}}{{\bf{\tilde d}}_i} + {{\bf{\dot d}}_i}.
\end{equation}
The above differential equations can be solved by multiplying both sides with $e^{\frac{1}{\varepsilon_i}t}$ and integrating over $\left[0,t\right]$ which yield
\begin{equation}\label{eq:dtildesolution}
{{\bf{\tilde d}}_i}\left( t \right) = {{\bf{\tilde d}}_i}\left( 0 \right){e^{ - \frac{1}{{{\varepsilon _i}}}t}} + \int_0^t  {{e^{ - \frac{1}{{{\varepsilon _i}}}\left( {t - \alpha } \right)}}{{{\bf{\dot d}}}_i}\left( \alpha   \right)} d\alpha.
\end{equation}
Writing \eqref{eq:dtildesolution} in a component-wise form and applying the absolute value operator $\left|\cdot\right|$ to both sides lead to
\begin{equation}\label{eq:triangleinequality}
\left| {\tilde d_i^j\left( t \right)} \right| \le \left| {\tilde d_i^j\left( 0 \right){e^{ - \frac{1}{{{\varepsilon _i}}}t}}} \right| + \left| {\int_0^t {{e^{ - \frac{1}{{{\varepsilon _i}}}\left( {t - \alpha } \right)}}\dot{d}_i^j\left( \alpha  \right)d\alpha } } \right|.
\end{equation}
According to Holder's inequality \cite{won2015high}, $\left| {\int_0^t {\chi\left( \tau  \right)d\tau } } \right| \le \int_0^t {\left| {\chi\left( \tau  \right)} \right|d\tau } $. 
Therefore,
\begin{equation}\label{eq:triangleinequality2}
    \left| {\tilde d_i^j\left( t \right)} \right| \le \left| {\tilde d_i^j\left( 0 \right){e^{ - \frac{1}{{{\varepsilon _i}}}t}}} \right| + \int_0^t {\left| {{e^{ - \frac{1}{{{\varepsilon _i}}}\left( {t - \alpha } \right)}}\dot d_i^j\left( \alpha  \right)} \right|d\alpha } .
\end{equation}
By integrating \eqref{eq:triangleinequality2} over $\left[0,t\right]$, we get
\begin{equation}\label{eq:integral1}
\begin{array}{l}
\int_0^t {\left| {\tilde d_i^j\left( \alpha  \right)} \right|d\alpha }  \le \int_0^t {\left| {\tilde d_i^j\left( 0 \right){e^{ - \frac{1}{{{\varepsilon _i}}}\alpha }}} \right|d\alpha } \\
\;\;\;\;\;\;\;\;\;\;\;\;\;\;\;\;\;\;\;\;\;\;\;\; + \int_0^t {\left( {\int_0^\beta  {{e^{ - \frac{1}{{{\varepsilon _i}}}\left( {\beta  - \alpha } \right)}}\dot d_i^j\left( \alpha  \right)d\alpha } } \right)d\beta } .
\end{array}
\end{equation}
The first term on the right-hand side of \eqref{eq:integral1} is bounded by ${\varepsilon _i}\left| {\tilde d_i^j(0)} \right| $.
The second term is equal to $\left\| {h\left(t\right) * \dot{d}_i^j\left(t\right)} \right\|$ where $h\left(t\right) = e^{-\frac{1}{\varepsilon_i}t}$ and $*$ is the convolution operator.
According to Young's convolution theorem \cite{wheeden2015measure}, $\left\| {h\left(t\right) * \dot d_i^j\left(t\right)} \right\| \le \left\| h\left(t\right) \right\|{\left\| \dot{d}_i^j\left(t\right) \right\|}$.
Also, from the definition of $\mathcal{L}_1$ norm, we have $\left\| {{h_i}\left( t \right)} \right\| = {\varepsilon _i}$.
Therefore,
\begin{equation}\label{eq:integral2}
\int_0^t {\left| {\tilde d_i^j\left( \alpha  \right)} \right|d\alpha }  \le {\varepsilon _i}\left| {\tilde d_i^j(0)} \right| + {\varepsilon _i}{\left\| {\dot d_i^j} \right\|}.
\end{equation}
Considering that the left-hand side of \eqref{eq:integral2} is the definition of $\mathcal{L}_1$ norm for $\tilde d_i^j$ and also using \eqref{eq:assumption} result in
\begin{equation}\label{eq:disturbanceEstBound}
{\left\| {\tilde d_i^j\left( t \right)} \right\|} \le {\varepsilon _i}\left| {\tilde d_i^j(0)} \right| + \delta_i^j ,
\end{equation}
which implies
\begin{equation}\label{eq:finaldtildebound}
\left\| {{{{\bf{\tilde d}}}_i}\left( t \right)} \right\| \le {\varepsilon _i}\left( {\left\| {{{{\bf{\tilde d}}}_i}\left( 0 \right)} \right\| + {{\boldsymbol{\delta }}_i}} \right).
\end{equation}
As $\varepsilon_i$ is a small constant, \eqref{eq:disturbanceEstBound} suggests that the disturbance estimation error is small and bounded for all time and this provides a theoretical verification for the convergence of the proposed HGDO.

\section{Controller Design}\label{se:controller}
\begin{figure}[t]
    \centering
    \includegraphics[trim={0cm 2cm 0cm 0cm}, clip, width = 0.8\linewidth]{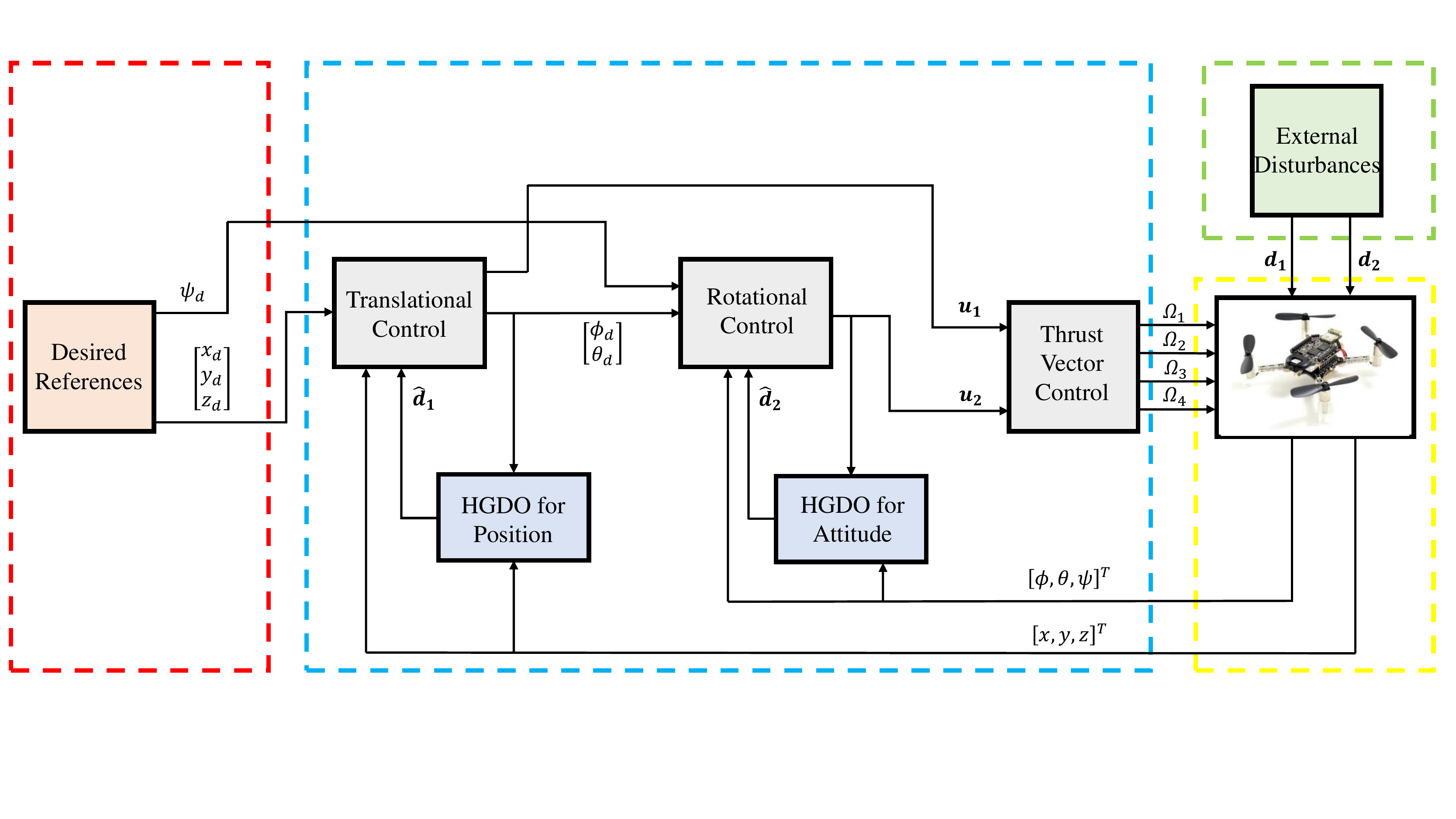}
    \caption{Overview of the control system architecture}
    \label{fig:block}
\end{figure}

Our objective in this section is to study how the proposed HGDO can be integrated into the Lyapunov-based control design.
We will study SMC design here as an example and establish stability results for the HGDO and the control law.
Note that we employ a cascaded structure to handle the underactuation of the quadrotor. As shown in Fig. \ref{fig:block}, the outer loop controls the transnational dynamics, while the inner loop tackles the rotational dynamics of the vehicle.

Let us start with the definition of tracking errors.
Given the desired trajectories $\left[{\bf{x}}_{1d}^T, {\bf{x}}_{3d}^T\right]^T$, we define the position tracking error as ${\bf{e}}_1 = {\bf{x}}_{1d} - {\bf{x}}_1$ and the attitude tracking error as ${\bf{e}}_2 = {\bf{x}}_{3d} - {\bf{x}}_3$.
Define sliding surfaces as follows
\begin{equation}\label{eq:s}
{{\bf{s}}_i} = {{\bf{\dot e}}_i} + {{\bf{\lambda }}_i}{{\bf{e}}_i},
\end{equation}
where ${{\boldsymbol{\lambda}}_{i}}\in {\mathbb{R}}^3, \;i=\left\{1,2\right\}$ are design parameters with positive components.
Let us now consider the following Lyapunov function candidate
\begin{equation}\label{eq:V}
V = \frac{1}{2}\left( {{\bf{s}}_1^T{{\bf{s}}_1} + {\bf{s}}_2^T{{\bf{s}}_2} + {\bf{\tilde d}}_1^T{{{\bf{\tilde d}}}_1} + {\bf{\tilde d}}_2^T{{{\bf{\tilde d}}}_2}} \right).
\end{equation}
Taking the derivative from both sides of \eqref{eq:V} and substituting \eqref{eq:estimationErrDyn2} and \eqref{eq:s} result in
\begin{equation}\label{eq:vdot}
{\dot V = {\bf{s}}_1^T{{{\bf{\dot s}}}_1} + {\bf{s}}_2^T{{{\bf{\dot s}}}_2} + {\bf{\tilde d}}_1^T{{{\bf{\dot{\tilde d}}}}_1} + {\bf{\tilde d}}_2^T{{{\bf{\dot{\tilde d}}}}_2}}.
\end{equation}
Using the tracking error definition and substituting system dynamics \eqref{eq:statespaceform} yield
\begin{equation}
\begin{array}{l}
\dot V = {\bf{s}}_1^T\left( {{{{\bf{\ddot x}}}_{1d}} + {\bf{g}} - {{\bf{u}}_{\bf{1}}} - {{\bf{d}}_1} + {{\boldsymbol{\lambda }}_1}{{{\bf{\dot e}}}_1}} \right)\\
\;\;\;\;\;\; + {\bf{s}}_2^T\left( {{{{\bf{\ddot x}}}_{3d}} - {{\bf{f}}_2}\left( {{{\bf{x}}_4}} \right) - {{\bf{u}}_2} - {{\bf{d}}_2} + {{\boldsymbol{\lambda }}_2}{{{\bf{\dot e}}}_2}} \right)\\
\;\;\;\;\;\; +{\bf{\tilde d}}_1^T{{{\bf{\dot{\tilde d}}}}_1} + {\bf{\tilde d}}_2^T{{{\bf{\dot{\tilde d}}}}_2}.
\end{array}
\end{equation}
Let us define the control laws as follows
\begin{equation}\label{eq:u}
\left\{ {\begin{array}{*{20}{l}}
{{{\bf{u}}_1} = {{{\bf{\ddot x}}}_{1d}} + {\bf{g}} - {{{\bf{\hat d}}}_1} + {{\bf{\lambda }}_1}{{{\bf{\dot e}}}_1} + {{\bf{k}}_1}{\mathop{\rm sgn}} \left( {{{\bf{s}}_1}} \right) + {{\bf{L}}_1}{{\bf{s}}_1},}\\
{{{\bf{u}}_2} = {{{\bf{\ddot x}}}_{3d}} - {{\bf{f}}_2}\left( {{{\bf{x}}_4}} \right) - {{{\bf{\hat d}}}_2} + {{\bf{\lambda }}_2}{{{\bf{\dot e}}}_2} + {{\bf{k}}_2}{\mathop{\rm sgn}} \left( {{{\bf{s}}_2}} \right) + {{\bf{L}}_2}{{\bf{s}}_2},}
\end{array}} \right.
\end{equation}
where $\rm{sgn}\left(\cdot\right)$ denotes the sign function and ${\bf{k}}_i \in \mathbb{R}^3$ and ${\bf{L}}_i \in \mathbb{R}^{3 \times 3}$ are design parameters.

Using \eqref{eq:states}, the individual components of ${{\bf{u}}_1}={[{u_x},{u_y},{u_z}]^T}$ can de determined as 
\begin{equation}\label{eq:u2}
\left\{ {\begin{array}{*{20}{l}}
{{u_x} = \frac{{{u_1}}}{m}(\cos \phi \sin \theta \cos \psi  + \sin \phi \sin \psi) ,}\\
{{u_y} = \frac{{{u_1}}}{m}(\cos \phi \sin \theta \sin \psi  - \sin \phi \cos \psi ),}\\
{{u_z} = \frac{{{u_1}}}{m}(\cos \phi \cos \theta). }
\end{array}} \right.
\end{equation}
Next, we determine the desired pitch angle $\phi_d$, desired roll angle $\theta_d$, and the input to the position controller, $u_1$, using \eqref{eq:u2}

\begin{equation}\label{eq:u3}
\left\{ {\begin{array}{*{20}{l}}
{{\theta _d} = \arctan [\frac{{{u_x}\cos {\psi _d} + {u_y}\sin {\psi _d}}}{{{u_z}}}],}\\
{{\phi _d} = \arctan [\cos {\theta _d}\frac{{{u_x}\sin {\psi _d} - {u_y}\cos {\psi _d}}}{{{u_z}}}],}\\
{{u_1} = \frac{{m{u_z}}}{{\cos {\phi _d}\cos {\theta _d}}}}.
\end{array}} \right.
\end{equation}


In order to ensure system stability, ${\bf{L}}_i$ must be positive-definite, and ${\bf{k}}_i$ must satisfy a criterion to be detailed shortly. Substituting \eqref{eq:u} in \eqref{eq:vdot} leads to
\begin{equation}\label{eq:vdot2}
\begin{array}{*{20}{l}}
{\dot V \le  - {\bf{s}}_1^T{{{\bf{\tilde d}}}_1} - {\bf{s}}_1^T{{\bf{k}}_1}{\mathop{\rm sgn}} \left( {{{\bf{s}}_1}} \right) - {\bf{s}}_1^T{{\bf{L}}_1}{{\bf{s}}_1}}\\
{\;\;\;\;\;\;\;\; - {\bf{s}}_2^T{{{\bf{\tilde d}}}_2} - {\bf{s}}_2^T{{\bf{k}}_2}{\mathop{\rm sgn}} \left( {{{\bf{s}}_2}} \right) - {\bf{s}}_2^T{{\bf{L}}_2}{{\bf{s}}_2}}\\
{\;\;\;\;\;\;\;\; + {\bf{\tilde d}}_1^T{{{\bf{\dot{\tilde d}}}}_1} + {\bf{\tilde d}}_2^T{{{\bf{\dot{\tilde d}}}}_2}.}
\end{array}
\end{equation}
For the last two terms in \eqref{eq:vdot2}, we use \eqref{eq:estimationErrDyn2} to write
\begin{equation}\label{eq:dterms}
    {\bf{\tilde d}}_1^T{{\bf{\dot{\tilde d}}}_1} + {\bf{\tilde d}}_2^T{{\bf{\dot{ \tilde d}}}_2} =  - \frac{1}{{{\varepsilon _1}}}{\bf{\tilde d}}_1^T{{\bf{\tilde d}}_1} + {\bf{\tilde d}}_1^T{{\bf{\dot d}}_1} - \frac{1}{{{\varepsilon _2}}}{\bf{\tilde d}}_2^T{{\bf{\tilde d}}_2} + {\bf{\tilde d}}_2^T{{\bf{\dot d}}_2}.
\end{equation}
For ${\bf{\tilde d}}_i^T{{{\bf{\dot d}}}_i}$ terms, we can use \eqref{eq:assumption} and \eqref{eq:finaldtildebound} to find upper bounds $\rho_i$ as follows
\begin{equation}\label{eq:dterms2}
{\bf{\tilde d}}_i^T{{\bf{\dot d}}_i} \le {\varepsilon _i}\left( {\left\| {{{{\bf{\tilde d}}}_i}\left( 0 \right)} \right\| + \left\| {{{\boldsymbol{\delta }}_i}} \right\|} \right)\left\| {{{\boldsymbol{\delta }}_i}} \right\| \le {\rho _i}
\end{equation}
 
By substituting \eqref{eq:estimationErrDyn2}, \eqref{eq:dterms}, and \eqref{eq:dterms2} in \eqref{eq:vdot2}, we get
\begin{equation}
\begin{array}{*{20}{l}}
{\dot V \le {\varepsilon _1}\left\| {{{\bf{s}}_1}} \right\|\left( {\left\| {{{{\bf{\tilde d}}}_1}\left( 0 \right)} \right\| + \left\| {{{\boldsymbol{\delta }}_1}} \right\| - \frac{1}{{{\varepsilon _1}}}{{\bf{k}}_1}} \right) - {\bf{s}}_1^T{{\bf{L}}_1}{{\bf{s}}_1}}\\
{\;\;\;\;\;\; + {\varepsilon _2}\left\| {{{\bf{s}}_2}} \right\|\left( {\left\| {{{{\bf{\tilde d}}}_2}\left( 0 \right)} \right\| + \left\| {{{\boldsymbol{\delta }}_2}} \right\| - \frac{1}{{{\varepsilon _2}}}{{\bf{k}}_2}} \right) - {\bf{s}}_2^T{{\bf{L}}_2}{{\bf{s}}_2}}\\
{\;\;\;\;\;\; - \frac{1}{{{\varepsilon _1}}}{\bf{\tilde d}}_1^T{{{\bf{\tilde d}}}_1} - \frac{1}{{{\varepsilon _2}}}{\bf{\tilde d}}_2^T{{{\bf{\tilde d}}}_2} + {\rho _1} + {\rho _2}.}
\end{array}
\end{equation}
Let us now choose ${\bf{k}}_i$ in a way that its components satisfy  
\begin{equation}\label{eq:kcondition}
k_i^j > {\varepsilon _i}\left( {\left|\tilde d_i^j(0)\right| + \delta _i^j} \right).
\end{equation}
Then, we can write
\begin{equation}\label{eq:vdot4}
\begin{array}{l}
\dot V \le  - {\bf{s}}_1^T{\lambda _{\max }}\left( {{{\bf{L}}_1}} \right){{\bf{s}}_1} - {\bf{s}}_2^T{\lambda _{\max }}\left( {{{\bf{L}}_2}} \right){{\bf{s}}_2}\\
\;\;\;\;\;\; - \frac{1}{{{\varepsilon _1}}}{\bf{\tilde d}}_1^T{{{\bf{\tilde d}}}_1} - \frac{1}{{{\varepsilon _2}}}{\bf{\tilde d}}_2^T{{{\bf{\tilde d}}}_2} + {\rho _1} + {\rho _2},
\end{array}
\end{equation}
where $\lambda_{max}\left(\cdot\right)$ denotes the largest eigenvalue.
We can now write
\begin{equation}\label{eq:vdot5}
    \dot V \le  - \kappa V + \rho,
\end{equation}
where $\kappa  = \min \left\{ {{\lambda _{\max }}\left( {{{\bf{L}}_1}} \right),{\lambda _{\max }}\left( {{{\bf{L}}_2}} \right),\frac{1}{{{\varepsilon _1}}},\frac{1}{{{\varepsilon _2}}}} \right\}$ and $\rho = \rho_1 + \rho_2$.
Taking steps similar to \eqref{eq:estimationErrDyn}-\eqref{eq:dtildesolution}, we get
\begin{equation}
V\left( t \right) \le {e^{ - \kappa t}}V\left( 0 \right) + \int_0^t {{e^{ - \kappa \left( {t - \alpha } \right)}}} \rho \left( \alpha  \right)d\alpha,
\end{equation}
which implies $V$ is bounded.
Subsequently, ${\bf{s}}_i$s are bounded.
This means that the tracking errors are bounded.
Therefore, with the control law \eqref{eq:u} and \eqref{eq:kcondition}, the vehicle trajectory will remain in an adjustable neighborhood of the desired trajectory.

One benefit of HGDO becomes clear in \eqref{eq:kcondition}. 
Note that $\varepsilon_i$ needs to be small to ensure fast disturbance estimation. 
This means that ${\varepsilon _i}\left( {\left| {\tilde d_i^j(0)} \right| + \delta _i^j} \right)$ have relatively small values, and thus, with small gains ${k}_i^j$ the stability condition \eqref{eq:kcondition} can be satisfied.
Small gains ${k}_i^j$ can potentially lead to smaller control efforts.

It is worth noting that the control laws in \eqref{eq:u} can lead to chattering due to the discontinuity of the sign function.
Therefore, for practical implementations, we will replace the sign function with saturation function ${\rm{sat}}\left(\frac{{\bf{s}}_i}{\mu}\right)$ where the slope of its linear portion is $\frac{1}{\mu}$. 
When $s_i^j > \mu$, the above stability results hold. 
Therefore, the system states will remain bounded even when the saturation function is used.

One major concern about HGO- and HGDO-based controllers is the peaking phenomenon that could result from very small $\varepsilon$ values.
In \cite{khalil2017high}, it is shown that passing the computed control signal from a saturation function can resolve the issue. As such, in our experiments presented in the next section, we will apply a saturation function to the computed signal to address the peaking phenomenon.
\section{Experiments}
This section presents the results of our simulation and laboratory experiments to evaluate the effectiveness of the proposed HGDO-based control. 
Our primary focus will be on laboratory experiments.
However, simulations are necessary to assess the accuracy of disturbance estimations; as it is otherwise challenging due to the difficulty of measuring the exact values of disturbances in practice.

{To highlight the benefits gained by HGDO, we compare our HGDO+SMC approach with the uncertainty and disturbance estimator method (UDE) \cite{talole2008model},  SMC-only, and also with one of the recent DO-based control methods \cite{kabiri20193d} that have tackled a similar problem.}

The vehicle under consideration throughout the experiments was a Crazyflie 2.1, flying in a controlled environment equipped with the lighthouse positioning system \cite{crazyflie}.
All the flight control computations were conducted on an external computer using MATLAB and transferred to the vehicle in real-time using Robot Operating System (ROS).

The parameter values for the vehicle, SMC, and HGDO used throughout simulations and real experiments are given in Tab. \ref{tab:params}. For tuning the SMC parameters, we used the genetic algorithm where the objective function was the integral of squared tracking errors. 
Once the SMC parameters were optimized, we tuned the HGDO parameters in a quick trial-and-error process.
To show the effect of $\varepsilon$ in HGDO performance, we present results for three different values of 0.01, 0.04, and 0.08.

\begin{table}[t]
\small
\caption{Parameter values used during experiments}
\label{tab:params}
\centering
\begin{tabular}{@{}lll@{}}
\toprule
& Parameter  & Value \\ \midrule
\multicolumn{3}{l}{\textit{a) Vehicle Parameters:}} \\
& $m$ & $28\;g$ \\
& ${\bf{J}} $ & ${\rm{diag}}\left( {1.4,1.4,2.17} \right) \times {10^{ - 5}}\;kg{m^2} $\\
& $k_T$ & $ 2.88 \times {10^{ - 8}}\;Ns^2 $ \\
& $k_Q$ & $ 7.24 \times {10^{ - 10}}\;Nms^2 $ \\
& $l$ & $92\;mm$ \\
\multicolumn{3}{l}{\textit{b) SMC Parameters:}} \\
 & ${{\boldsymbol{\lambda }}_1} $ & $ \left[0.3580, 0.5058, 0.3405\right]^T$ \\
 & ${{\boldsymbol{\lambda }}_2}$ & $\left[0.3580, 0.5058, 0.3405\right]^T$ \\
 & ${{\bf{k}}_1}$ & $\left[-5.2608, -5.0176, -5.4351\right]^T$ \\
 & ${{\bf{k}}_2}$ & $ [-8.0568, -13.6547, -1.8914]^T$ \\
 & ${{\bf{L}}_1}$ & ${\rm{diag}}\left( { - 2.6304, - 2.5088, - 2.7176} \right)$ \\
 & ${{\bf{L}}_2}$ & ${\rm{diag}}\left( { - 4.0284, - 6.8274, - 0.9457} \right)$ \\
\multicolumn{3}{l}{\textit{c) HGDO Parameters:}} \\
 & $\varepsilon_1$ & $0.01$ \\
 & $\varepsilon_2$ & $ 0.01$ \\
 \bottomrule
\end{tabular}
\end{table}

\subsection{Simulations}
The desired trajectories of the quadrotor were chosen as ${{\bf{x}}_{1d}} = {[ {0.5\sin \left( {\frac{{2\pi t }}{{40}}} \right),\sin \left( {\frac{{2\pi t }}{{40}}} \right)\cos \left( {\frac{{2\pi t }}{{40}}} \right),0.5} ]^T}$ and ${{\bf{x}}_{3d}} = {\left[ {0,0,0} \right]^T}$.
{For external disturbances, we used a sinusoidal disturbance with the maximum frequency of 4 Hz as
$d = 0.05[\sin(8\pi t)+\sin \left( {2.5\pi \left( {t - 3} \right)} \right) + 1.5\sin \left( {2\pi \left( {t + 7} \right)} \right) + 2\sin \left( {0.4\pi \left( {t - 9} \right)} \right) + \sin \left( {0.2\pi t} \right) + 0.5\sin \left( {0.08\pi \left( {t + 1} \right)} \right) + \sin \left( {0.07\pi \left( {t + 1.5} \right)} \right) + 0.5\sin \left( {0.05\pi \left( {t + 2} \right)} \right) + 4]$. This disturbance is taken from \cite{kabiri20193d}, with the addition of a $\sin(8\pi t)$ term to account for relatively higher frequency disturbances.
We also used the Dryden wind turbulence model \cite{beal1993digital}, a widely recognized model for its applicability in capturing real-world disturbances.


\begin{figure}[t]
    \centering
    \includegraphics[trim={0cm 0cm 0cm 0cm}, clip, width = 0.75\linewidth]{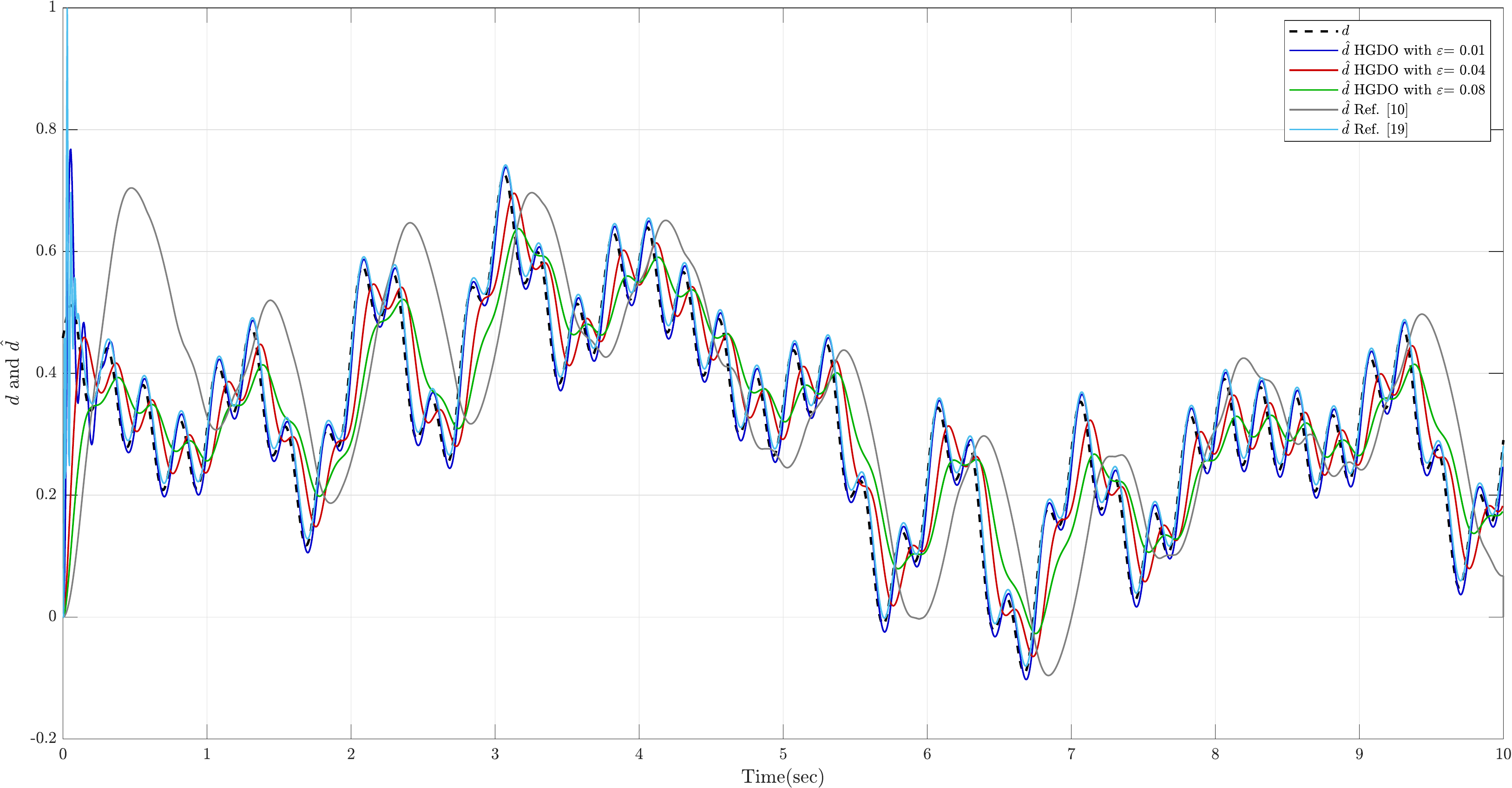}
    \caption{Sinusoidal disturbance with the maximum frequency of 4 Hz and its estimation}
    \label{fig:SimSin}
\end{figure}


    \begin{figure}[t]
    \centering
    \includegraphics[trim={0cm 0cm 0cm 0cm}, clip, width = 0.75\linewidth]{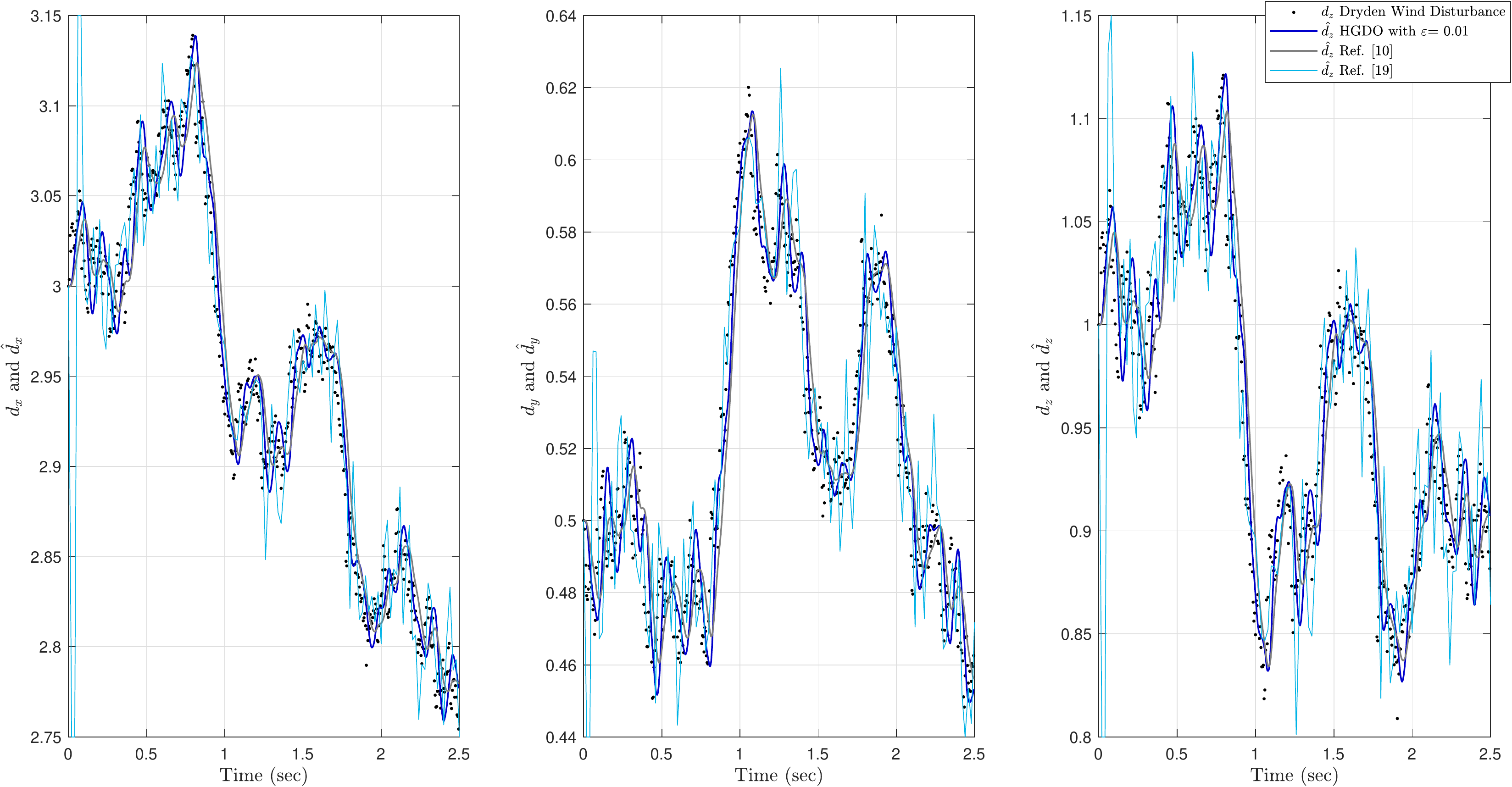}
    \caption{Dryden wind disturbance and its estimation}
    \label{fig:SimDryden}
\end{figure}

Figures \ref{fig:SimSin} and \ref{fig:SimDryden} compare the actual disturbance and the estimated values using HGDO and the DOs given in \cite{kabiri20193d} and \cite{talole2008model}.
In Fig. \ref{fig:SimSin}, among HGO results with different gains, the best disturbance estimation results correspond to $\varepsilon = 0.01$, as expected.
The differences are clear in both convergence speed and estimation error. It is easy to verify that the value of $\varepsilon$ directly influences the speed and accuracy of estimation, providing an easy way to calibrate the DO.

Comparing the HGDO with the method in \cite{kabiri20193d}, it is evident that even with $\varepsilon=0.08$, the HGDO exhibits higher convergence speed.
In terms of estimation accuracy, the difference between the method in \cite{kabiri20193d} and HGDO with $\varepsilon=0.08$ becomes insignificant over time; however, when $\varepsilon$ is set to 0.01, HGDO has a clear advantage.  

The UDE method described in \cite{talole2008model} exhibits a considerable overshoot at the start of the simulation. While this overshoot signifies the method's ability to address disturbances promptly, it could present practical difficulties, particularly in situations requiring precise and immediate adherence to a set trajectory without initial deviations. 
Once the transient behavior is passed, the UDE's estimation accuracy is comparable to HGDO with $\varepsilon = 0.01$; however, the HGDO has a smaller overshoot at the initial phase.

Concerning Fig. \ref{fig:SimDryden}, although the disturbance generated by the Dryden model is stochastic; the disturbance estimates have converged to the actual disturbance values only after a short transient time; however, the HGDO estimates are much closer to the actual disturbances compared to the two alternative methods.
Such a fast and accurate disturbance estimation presents a significant advantage in disturbance compensation and robust trajectory tracking.



\begin{figure}[t]
    \centering
    \includegraphics[trim={0cm 0cm 0cm 0cm}, clip, width = 0.75\linewidth]{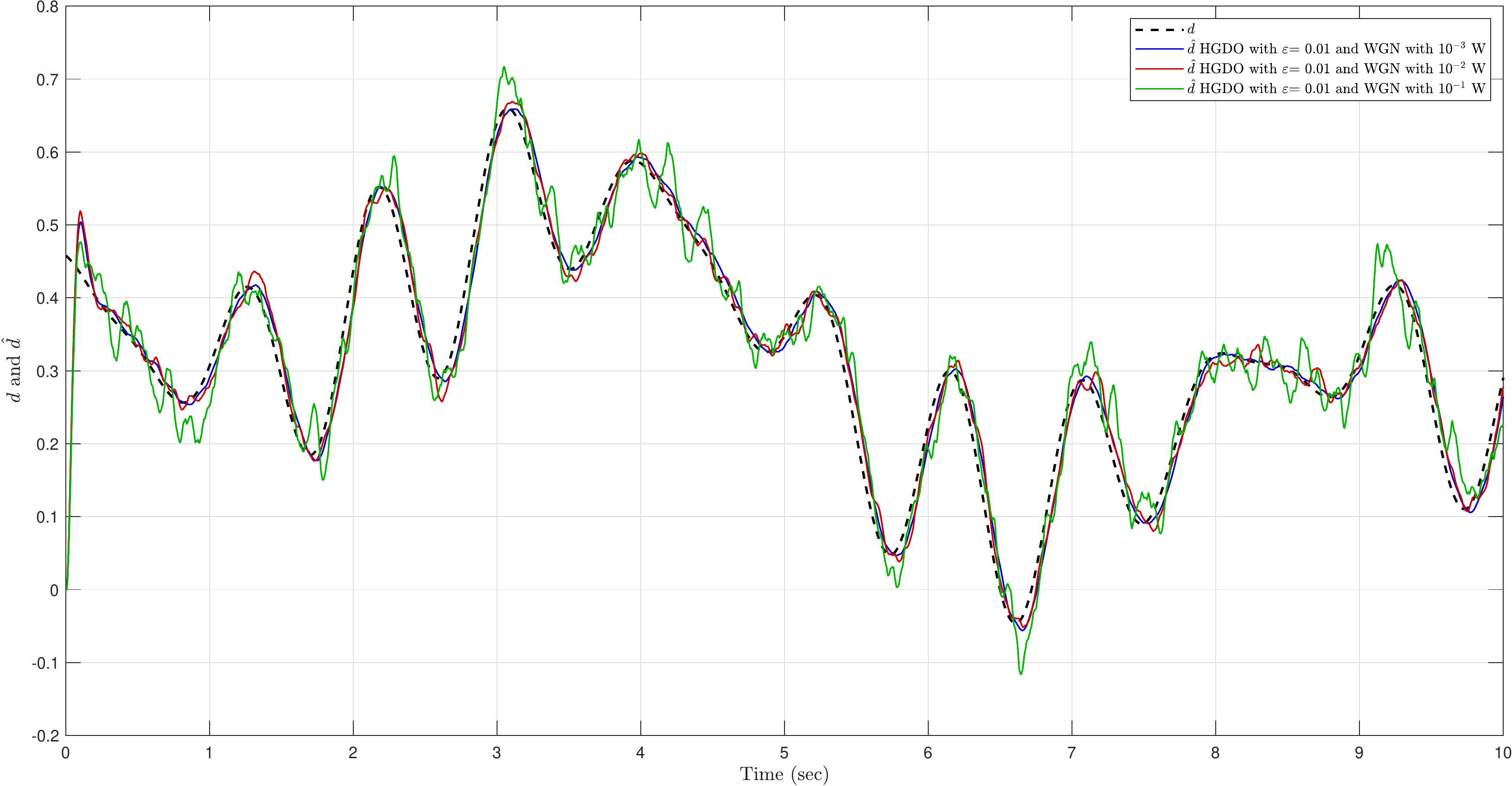}
    \caption{Sinusoidal disturbance and its estimation in the presence of noise. }
    \label{fig:noise}
\end{figure} 
{To investigate the HGDO performance in the presence of noise, we used white Gaussian measurement noise in our simulations, conducting three trials, each with a  different noise power (0.001, 0.01, and 0.1 W) as shown in Fig. \ref{fig:noise}.
HGOs are known to be sensitive to measurement noise, and this is evident in simulation results, especially for the highest noise power considered.
However, the estimation results are still reasonable, with an accuracy comparable to our benchmark methods with no measurement noise.
However, if the jitter in HGDO estimations is deemed to be problematic for a certain application, there exists a large body of literature on dealing with measurement noise for HGOs, e.g., by switching between two gains \cite{ahrens2009high}, gain adaptation \cite{sanfelice2011performance}, or integration with Kalman filter \cite{boizot2010adaptive}.
Note that there exists a formal mathematical guarantee for the boundedness of HGO estimation errors in the presence of bounded measurement noise, detailed in Theorem 8.1 of \cite{khalil2017high}.}

\begin{figure}[t]
    \centering
    \includegraphics[trim={0cm 0cm 0cm 0cm}, clip, scale = 0.36]{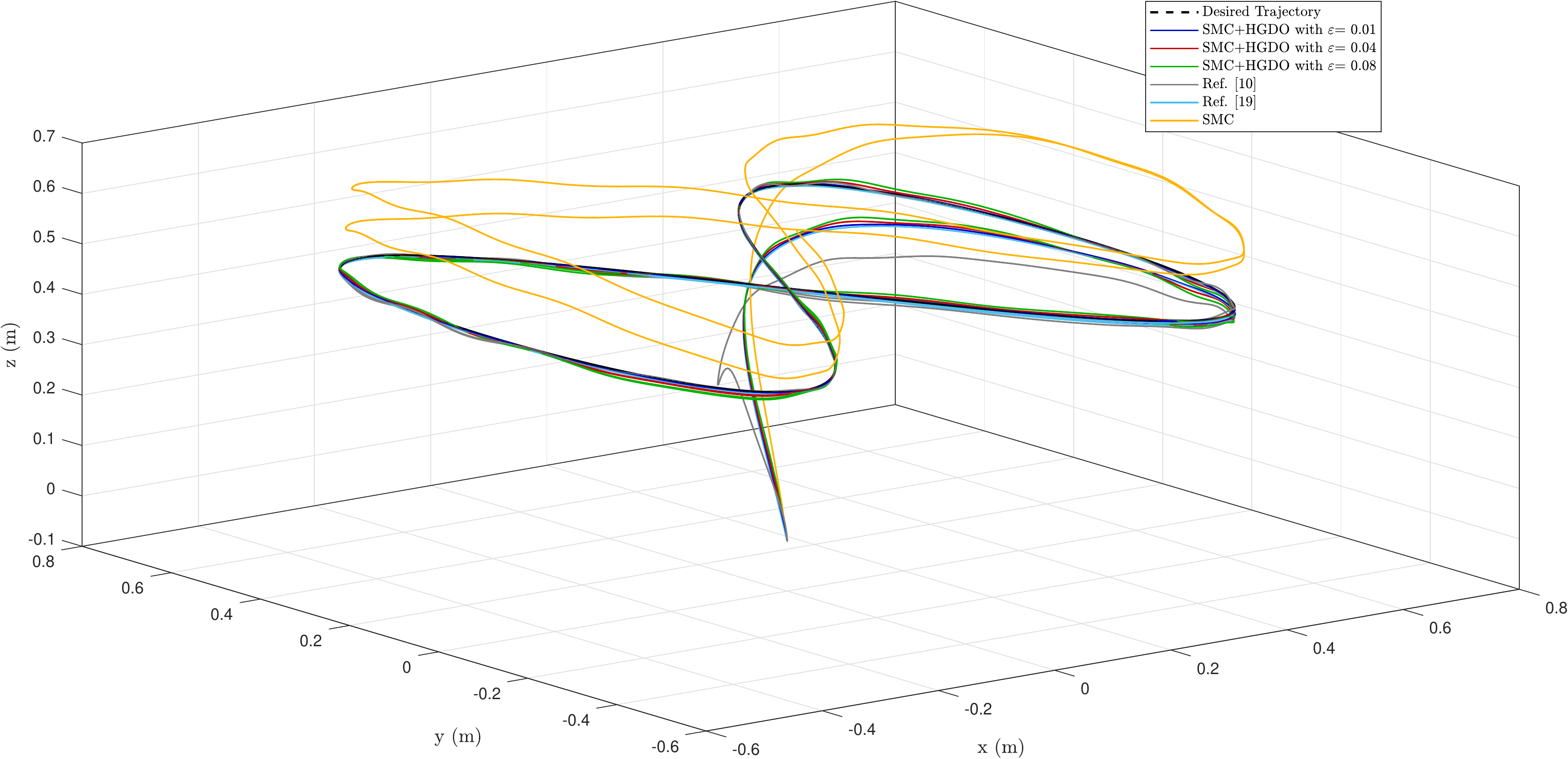}
    \caption{3D plot of the vehicle trajectory compared to the desired trajectory in the simulation study.}
    \label{fig:3D_trajectory_sim}
    \centering
    \includegraphics[trim={0cm 0cm 0cm 0cm}, clip, width = 0.75\linewidth]{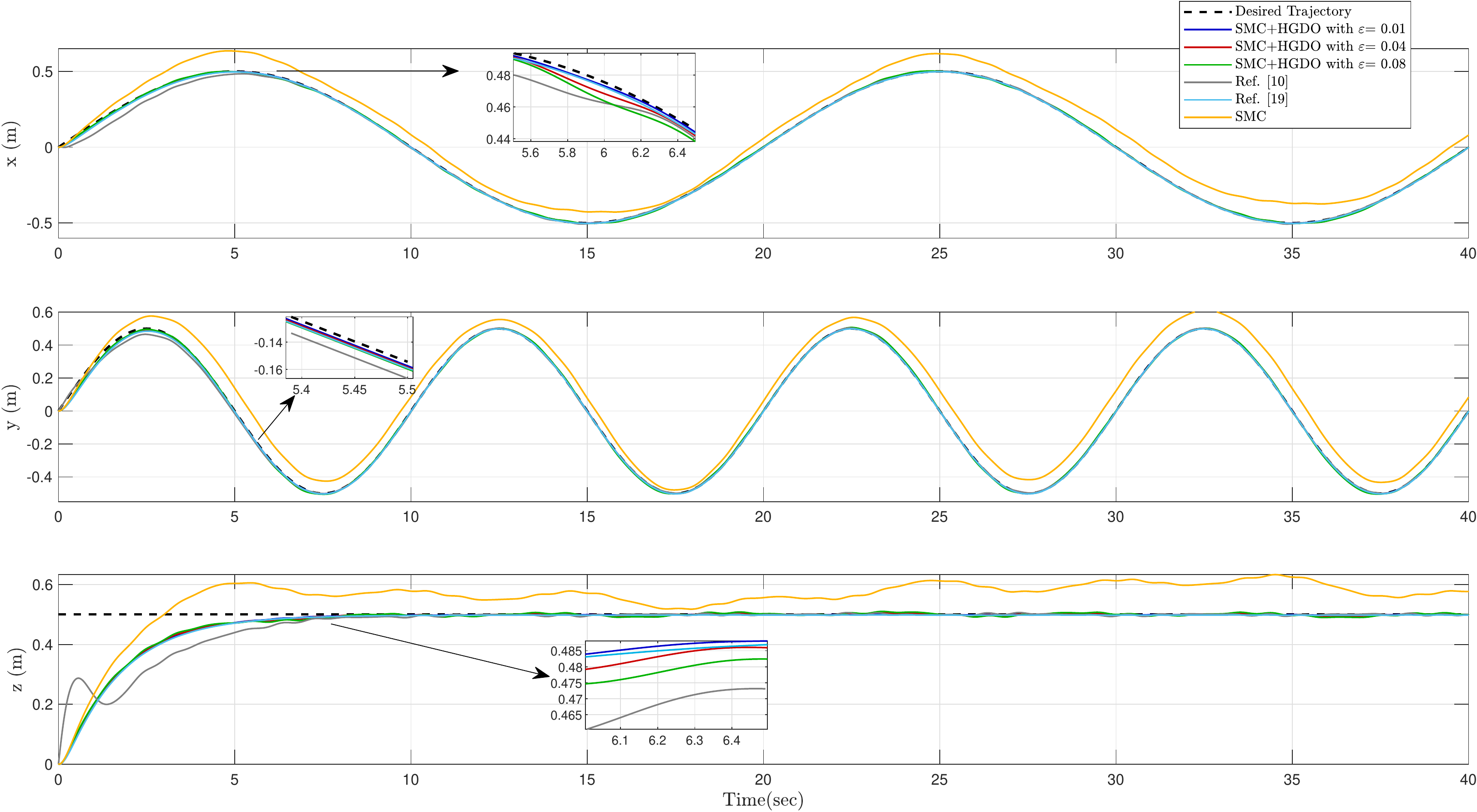}
    \caption{Desired and actual vehicle position trajectories in the simulation study.}
    \label{fig:position_tracking_sim}
\end{figure}
Figure \ref{fig:3D_trajectory_sim} depicts the 3D plot of the vehicle trajectory versus the desired trajectory in the time interval $ \left[0,40 s \right]$. 
Note that, in each trial, the vehicle completes the lemniscate figure twice.
The largest tracking error is associated with the SMC-only controller.
  When DO is introduced, the trajectory tracking is much improved; however, the convergence to the desired trajectory is noticeably slower with the DOs in \cite{kabiri20193d} and \cite{talole2008model}.
This likely stems from the slower convergence of disturbance estimations with these methods compared to the HGDO, as confirmed in Figs. \ref{fig:SimSin} and \ref{fig:SimDryden}.

Figure \ref{fig:position_tracking_sim} presents a closer look at the vehicle position.
The magnified windows provide a means to compare the magnitude of tracking errors.
Notably, HGDO with $\varepsilon=0.01$ exhibits a remarkably near-zero tracking error.
While tracking errors degrade by the increase in $\varepsilon$, the worst HGDO-based results still have faster convergence and similar tracking errors compared to the three alternative methods.
\subsection{Laboratory Experiments}

\subsubsection*{Scenario 1: Tracking a lemniscate trajectory}
\begin{figure}[t]
    \centering
    \includegraphics[trim={0cm 0cm 0cm 0cm}, clip, width = 0.75\linewidth]{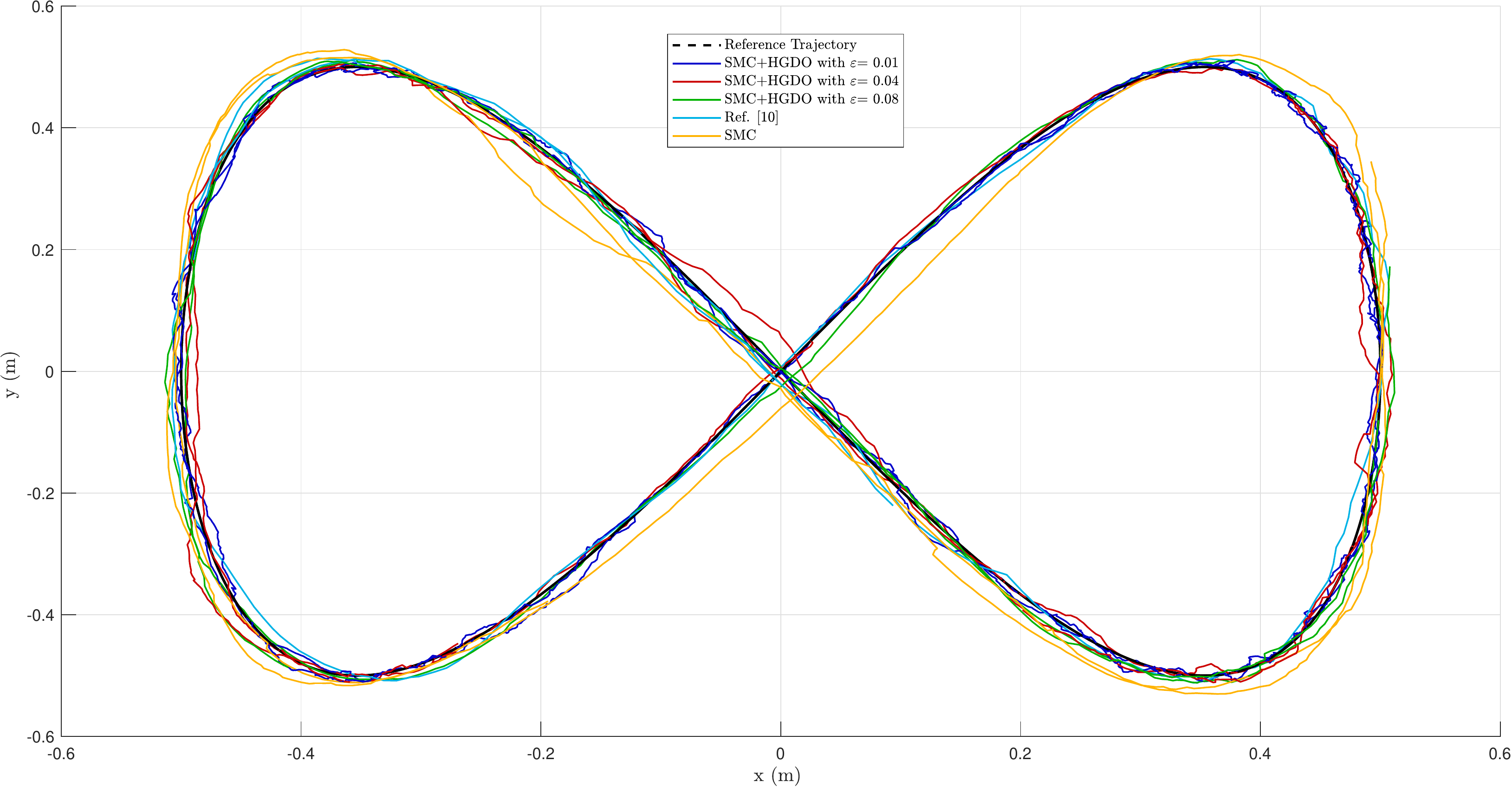}
    \caption{2D plot of the vehicle trajectory in comparison with the desired trajectory in tracking a lemniscate trajectory.}
    \label{fig:3D_trajectory_im}
\end{figure}
In this scenario, the vehicle tracks a lemniscate path similar to the one mentioned in simulations. An external fan generates a wind disturbance with a speed of $4\;km/h$ in the vicinity of the path, and the vehicle undergoes a non-uniform disturbance as it completes the path. 
Figure \ref{fig:3D_trajectory_im} presents the vehicle trajectory for the different methods.
Note that the vehicle completes the path twice to have a rough assessment of the repeatability of the results.
A closer look at the position states and the tracking error in Figs. \ref{fig:xyz_trajectory_im} and \ref{fig:error_position_im} highlight the superiority of HGDO+SMC, both in terms of convergence speed and tracking error.

{We also study the vehicle's attitude in this scenario, targeting a desired trajectory of $\left[\psi, \phi, \theta\right]^T=\left[0, 0, 0\right]^T$.
Comparing HGDO+SMC results with SMC-only and \cite{kabiri20193d} in Fig. \ref{fig:Euler_angles_tracking}, it is clear that HGDO+SMC presents outperforms the alternative methods.}

{To provide a quantitative assessment of the performance of the controllers, we tabulate the root mean square (RMS) of position and attitude tracking errors of the different control strategies in Tab. \ref{tab:trajectory}.
First, the SMC-only results have the highest error values, highlighting the benefits of adding a DO to achieve higher tracking accuracy. 
Second, comparing DO-based results shows that all HGDO+SMC results, even with $\varepsilon = 0.08$, exhibit smaller errors compared to \cite{kabiri20193d}.
As expected, the HGDO with $\varepsilon = 0.01$ has a clear advantage over all the other alternatives.}

\begin{figure}[H]
    \centering
    \includegraphics[trim={0cm 0cm 0cm 0cm}, clip, width = 0.75\linewidth]{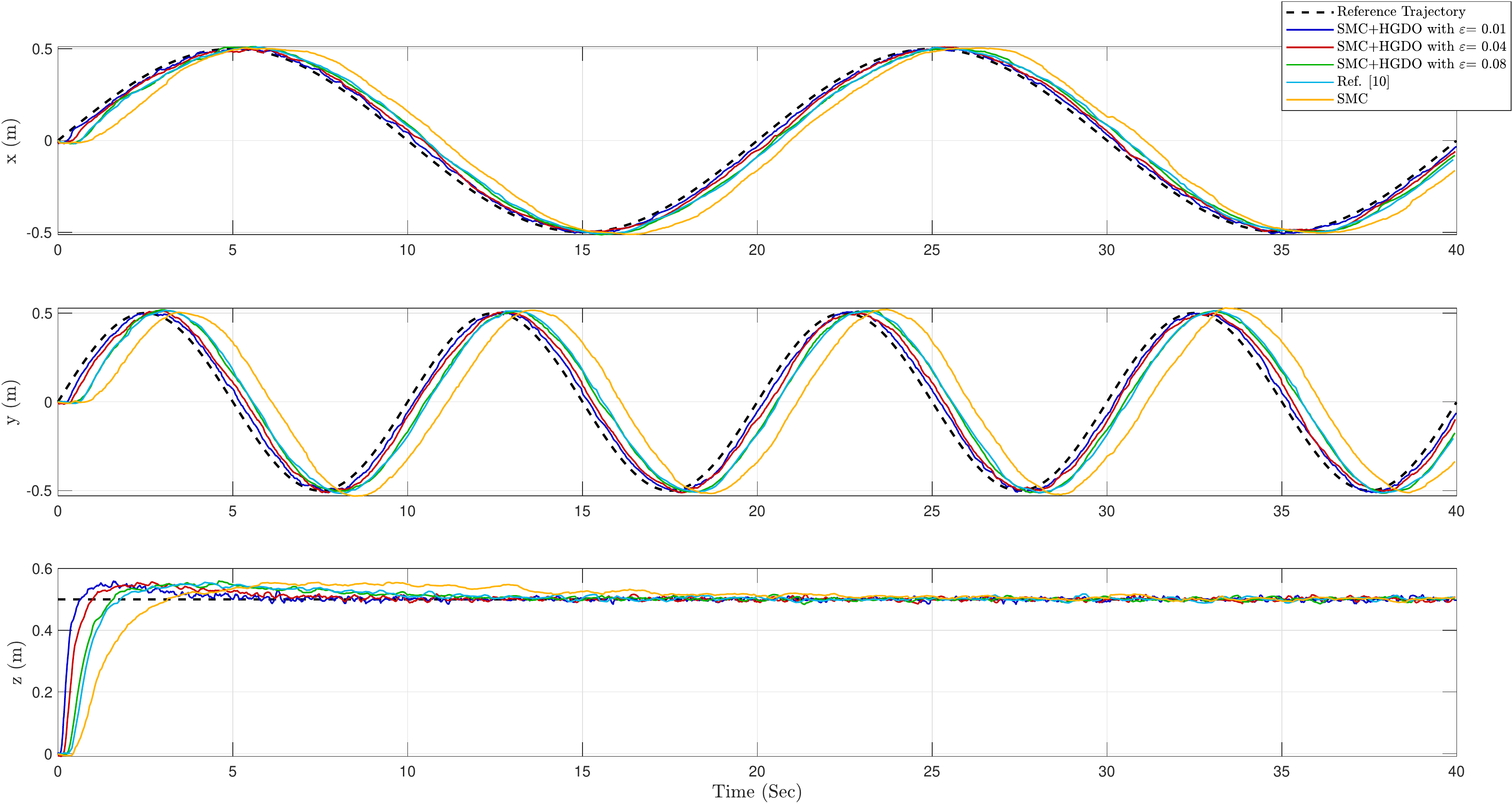}
    \caption{Desired and actual vehicle position trajectories in tracking a lemniscate trajectory scenario.}
    \label{fig:xyz_trajectory_im}
    \centering
    \includegraphics[trim={0cm 0cm 0cm 0cm}, clip, width = 0.75\linewidth]{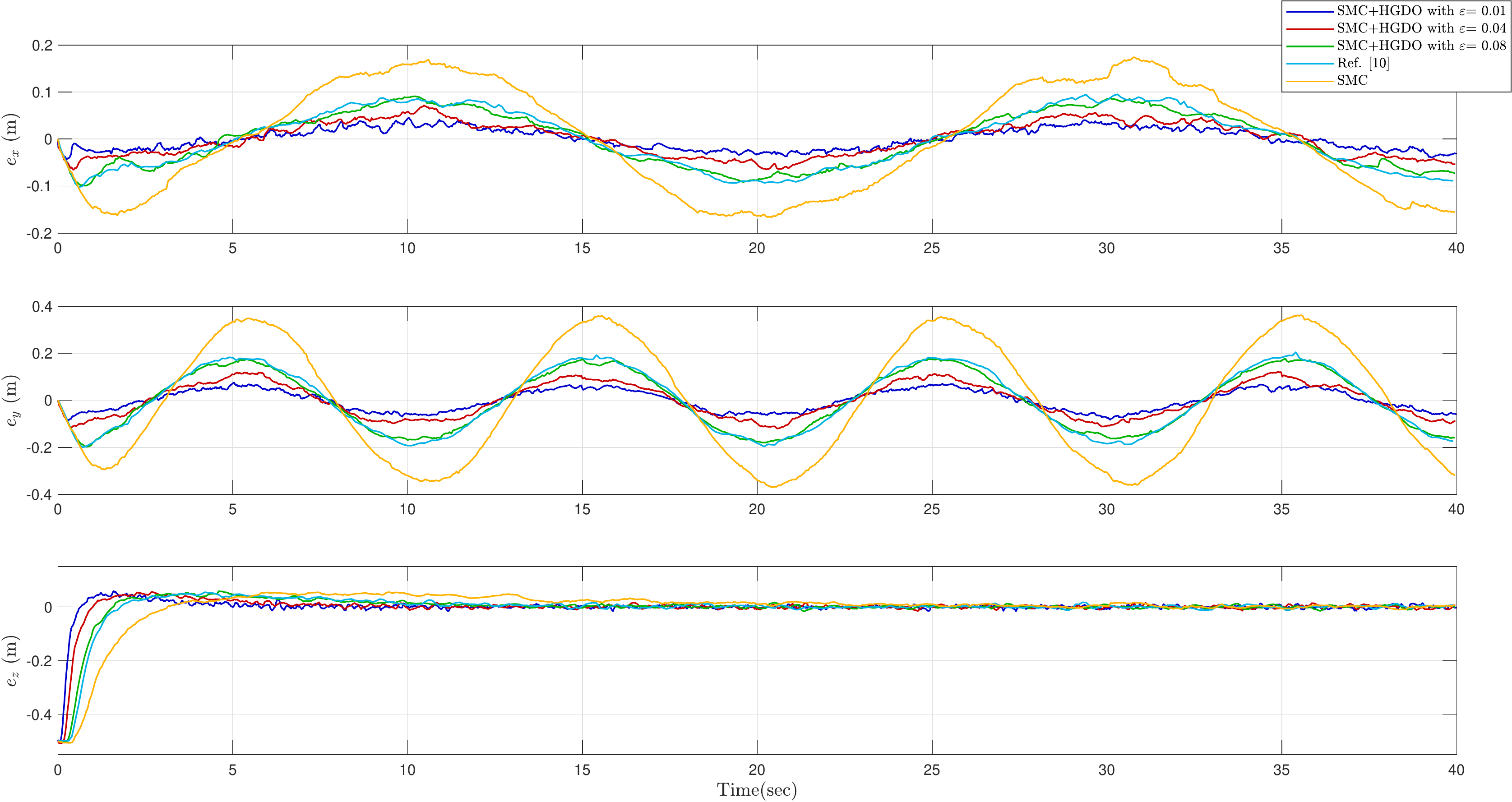}
    \caption{Position tracking error in tracking a lemniscate trajectory scenario.}
    \label{fig:error_position_im}
    \centering
    \includegraphics[trim={0cm 0cm 0cm 0cm}, clip, width = 0.75\linewidth]{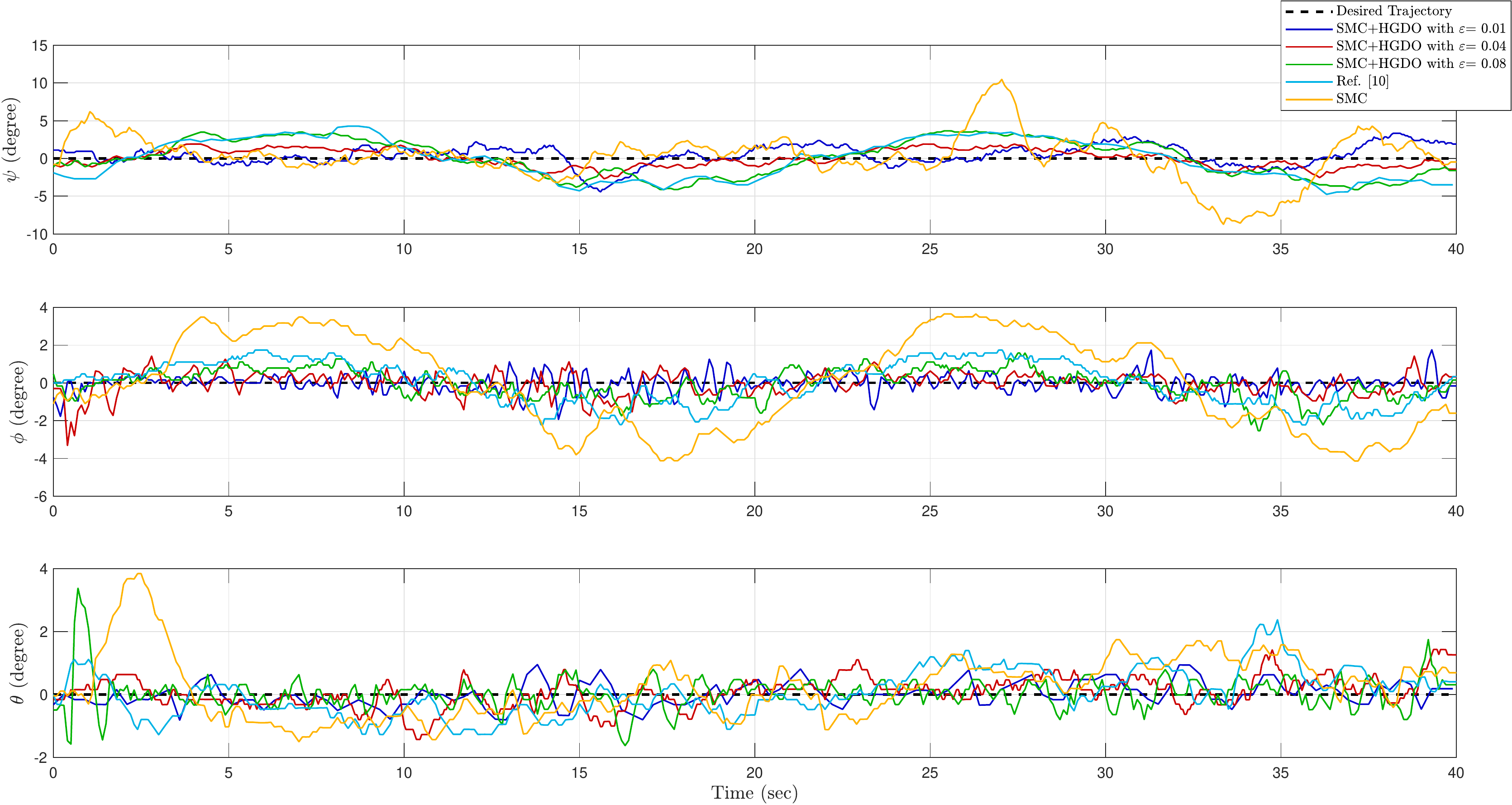}
    \caption{Desired and actual vehicle attitude in tracking a lemniscate trajectory scenario.}
    \label{fig:Euler_angles_tracking}
\end{figure}

\begin{table}[t]
  \caption{Root mean square of position and attitude tracking errors in tracking a lemniscate trajectory.}
\label{tab:trajectory}
  \resizebox{\textwidth}{!}{%
\begin{tabular}{@{}llllll@{}}
\toprule     
      Parameter & SMC+HGDO with $\varepsilon$= 0.01  & SMC+HGDO with $\varepsilon$= 0.04 & SMC+HGDO with $\varepsilon$= 0.08 & Ref. [10] & SMC \\
      \midrule
      $x$ & 0.022 & 0.035 & 0.056 & 0.060 & 0.111 \\
      $y$ & 0.044 & 0.070 & 0.119 & 0.128 & 0.242 \\
      $z$ & 0.037 & 0.050 & 0.064 & 0.072 & 0.018 \\
      $\psi$ & 0.021 & 0.032 & 0.041 & 0.044 & 0.054 \\     
      $\phi$ & 0.008 & 0.010 & 0.013 &0.019 & 0.041 \\     
      $\theta$ & 0.006 & 0.008 & 0.009 & 0.012 & 0.018 \\
      \bottomrule
    \end{tabular}
  }
\end{table}

\subsubsection*{Scenario 2: Ground effect}
In this scenario, we conducted a comprehensive evaluation of the vehicle's performance in close proximity to the ground, maintaining a minimal altitude of just 10 cm while executing a lemniscate trajectory, with the objective of ensuring that the vehicle's attitude remains stable and near zero throughout the flight. 
We did not use an external fan in this case to focus on another form of external disturbance that exists due to the airflow distortion near the ground known as the ground effect.

Figures \ref{fig:3D_trajectory} - \ref{fig:error_position_gr} illustrate the results of trajectory tracking for position, while Fig. \ref{fig:Euler_angles_ground} depicts the results for attitude tracking. Once again, these figures confirm the superior performance achieved through the use of HGDO+SMC.
Fig. \ref{fig:dis_estimation_gr} presents the real-time estimation of disturbances encountered during flight including the ground effect. 
Interestingly, the disturbance estimate along the $z$-axis is relatively larger, which can be explained by the presence of the ground effect.

{Table \ref{tab:ground} compares the RMS values of position and attitude tracking errors for different control strategies in this scenario, again, showing the superiority of HGDO+SMC to other methods.}

\begin{figure}[H]
    \centering
    \includegraphics[trim={0cm 0cm 0cm 0cm}, clip, width = 0.75\linewidth]{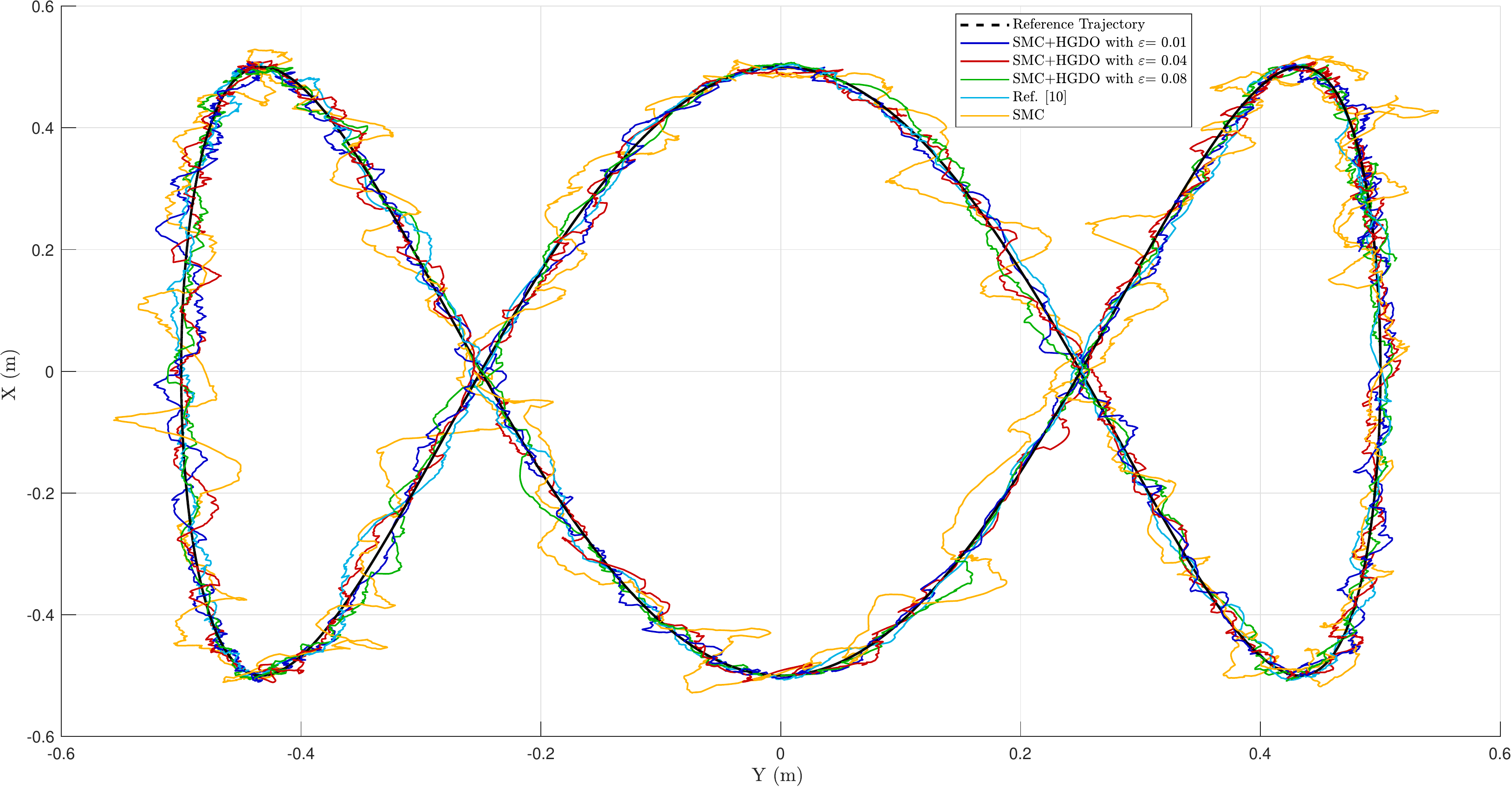}
    \caption{2D plot of the vehicle trajectory in comparison with the desired trajectory in the ground effect scenario.}
    \label{fig:3D_trajectory}
    \centering
    \includegraphics[trim={0cm 0cm 0cm 0cm}, clip, width = 0.75\linewidth]{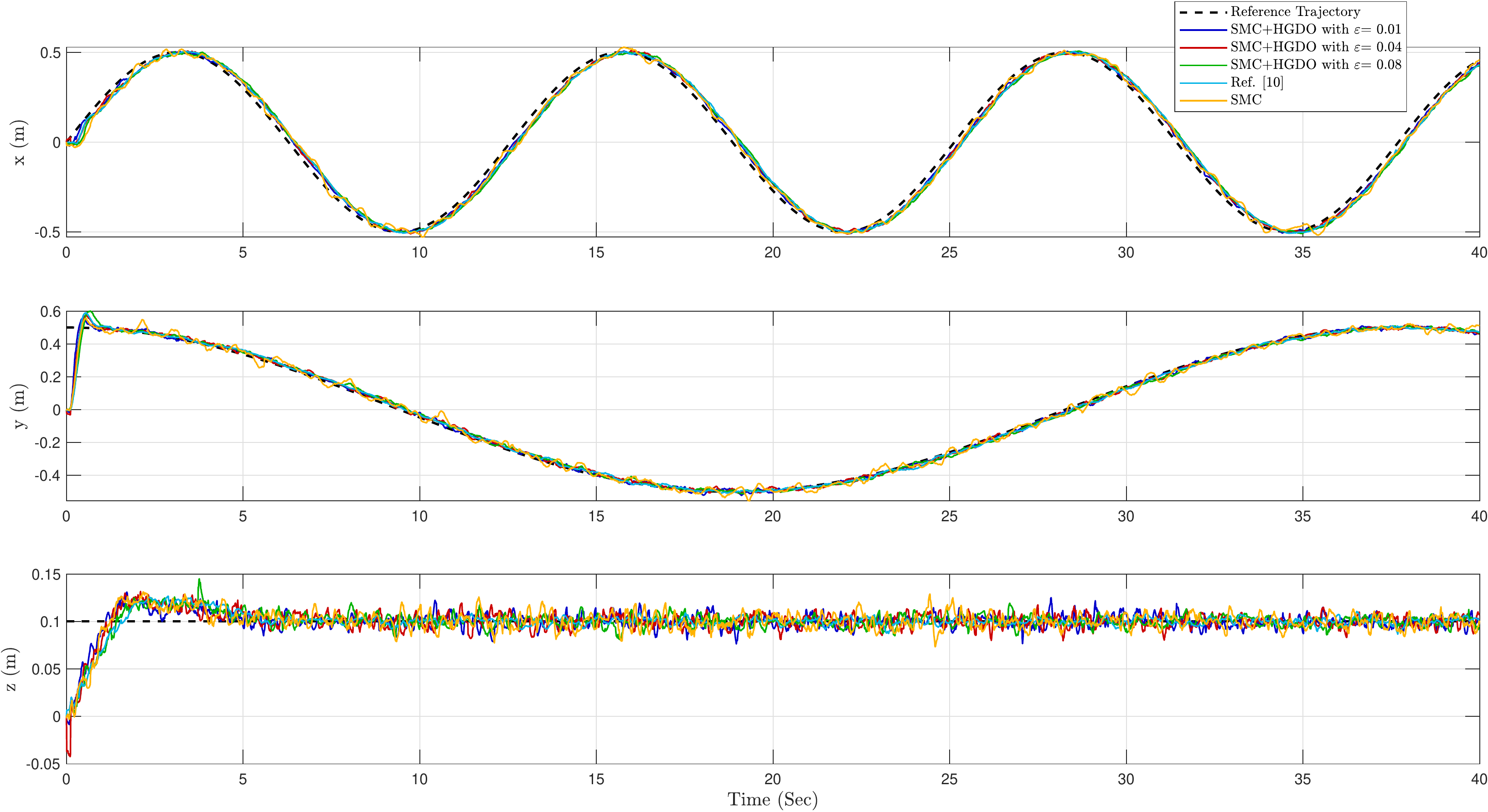}
    \caption{Desired and actual vehicle position trajectories in the ground effect scenario.}
    \label{fig:xyz_trajectory_gr}
    \centering
    \includegraphics[trim={0cm 0cm 0cm 0cm}, clip, width = 0.75\linewidth]{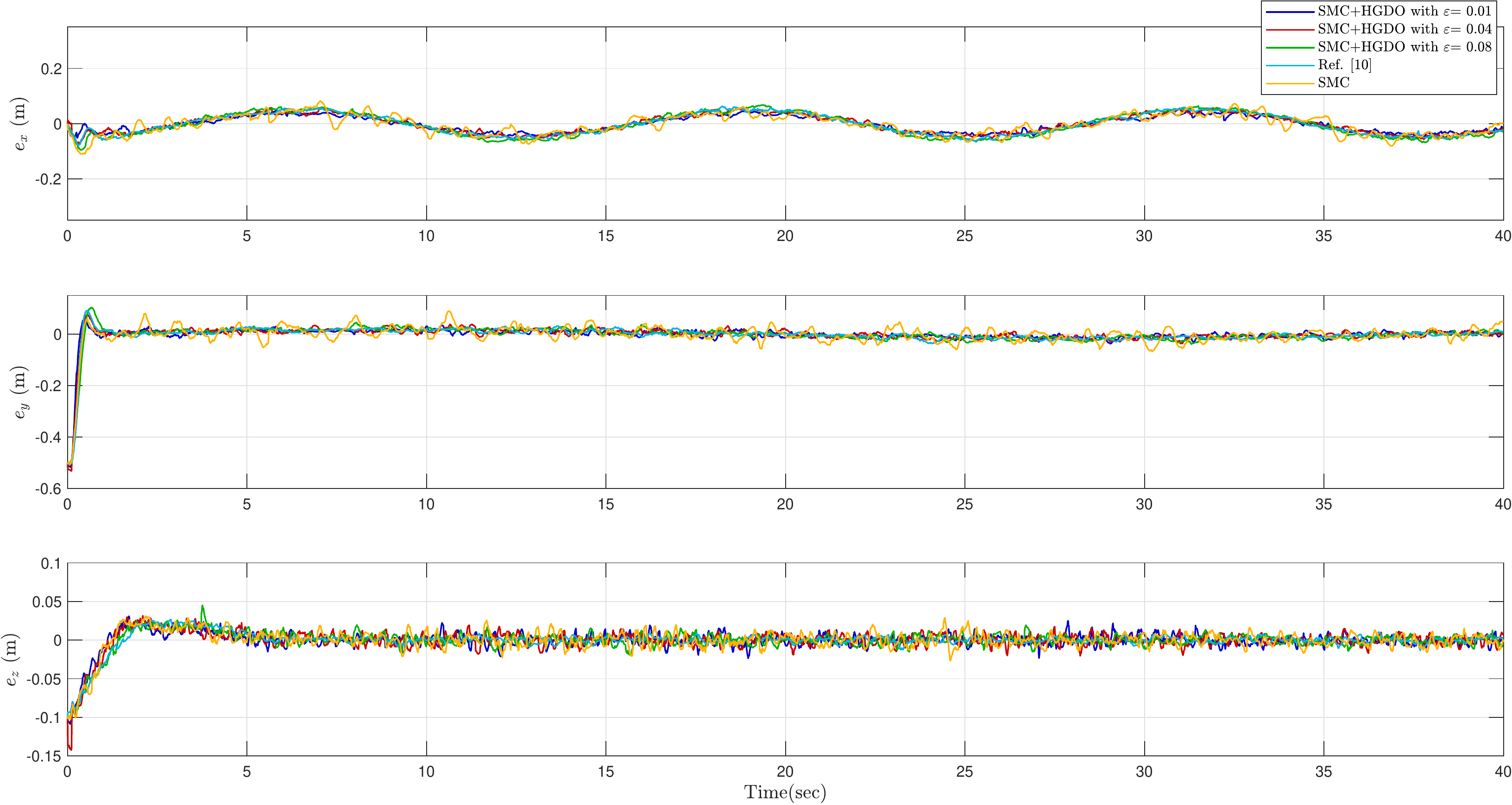}
    \caption{Position tracking error in the ground effect scenario.}
    \label{fig:error_position_gr}
\end{figure}
\begin{figure}[H]
    \centering
    \includegraphics[trim={0cm 0cm 0cm 0cm}, clip, width = 0.75\linewidth]{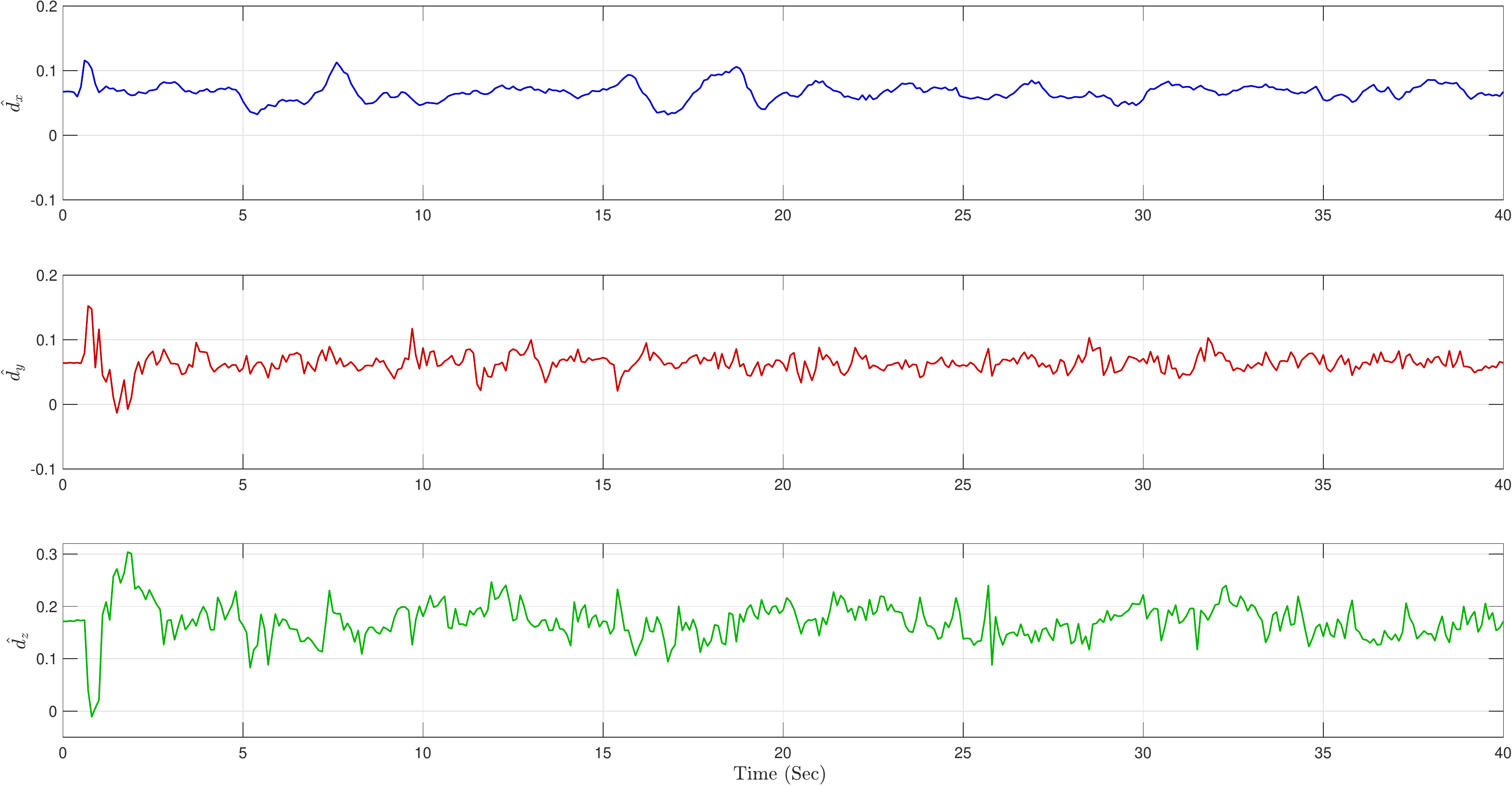}
    \caption{Disturbance estimates in the ground effect scenario.}
    \label{fig:dis_estimation_gr}
    \centering
    \includegraphics[trim={0cm 0cm 0cm 0cm}, clip, width = 0.75\linewidth]{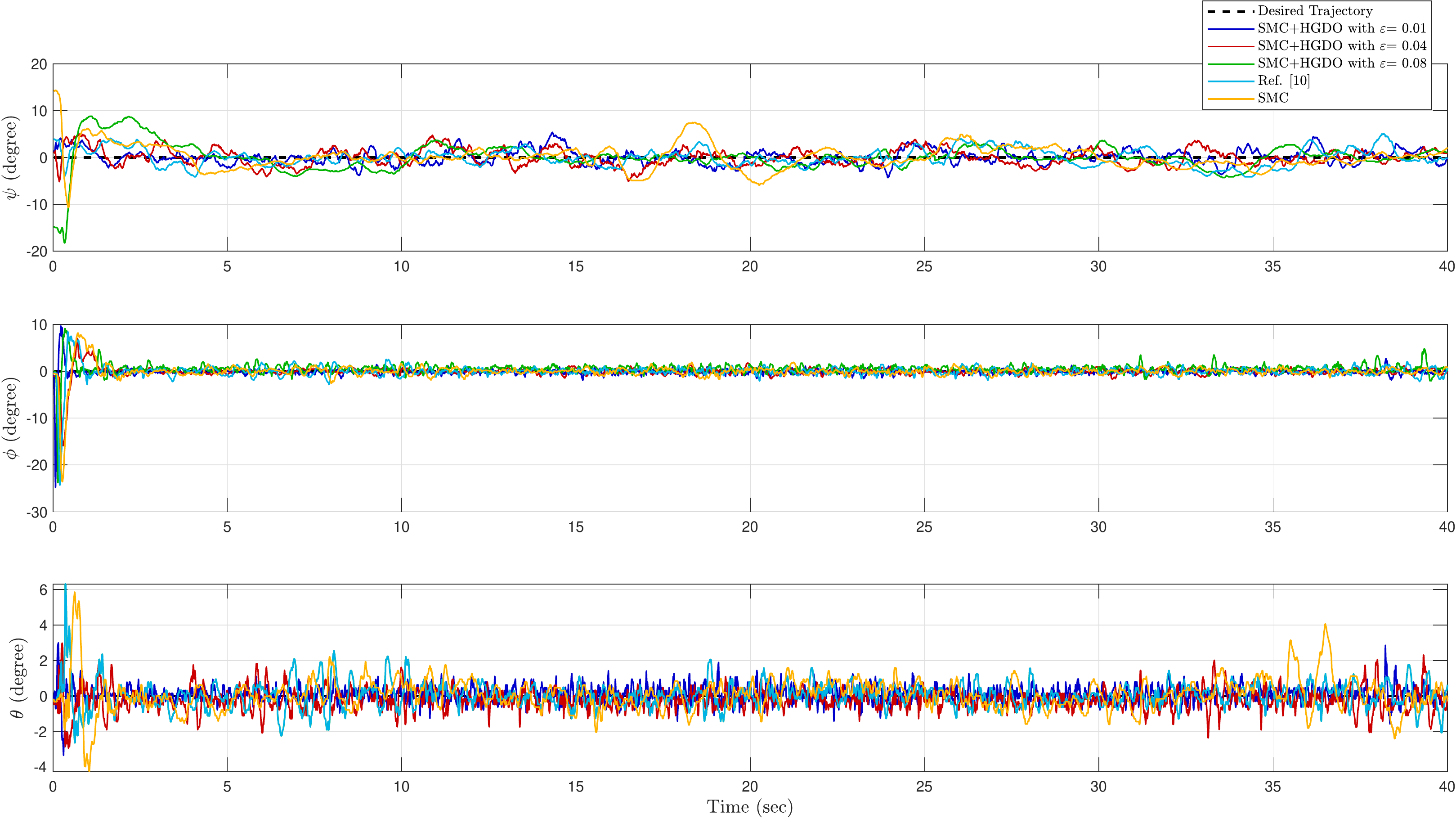}
    \caption{Desired and actual vehicle attitude in the ground effect scenario.}
    \label{fig:Euler_angles_ground}
\end{figure}

\begin{table}[H]
  \caption{Root mean square of position and attitude tracking errors in the ground effect scenario.}
  \label{tab:ground}
    \resizebox{\textwidth}{!}{%
\begin{tabular}{@{}llllll@{}}
\toprule  
      Parameter & SMC+HGDO with $\varepsilon$= 0.01  & SMC+HGDO with $\varepsilon$= 0.04 & SMC+HGDO with $\varepsilon$= 0.08 & Ref. [10] & SMC \\ \midrule
      $x$ & 0.028 & 0.033 & 0.036 & 0.037 & 0.042 \\
      $y$ & 0.037 & 0.041 & 0.043 &0.042 & 0.044 \\
      $z$ & 0.012 & 0.013 & 0.014 & 0.013 & 0.014 \\
      $\psi$ & 0.026 & 0.030 & 0.031 & 0.031 & 0.044 \\
      $\phi$ & 0.016 & 0.020 & 0.023 &0.024 & 0.028 \\
      $\theta$ & 0.007 & 0.010 & 0.012 & 0.013 & 0.016 \\ \bottomrule
    \end{tabular}
    }
\end{table}

\subsubsection*{Scenario 3: {Hovering control}}
In this scenario, the vehicle takes off from the origin, flies to point $\left[0.5,0.5,0.5\right]^T$, and hovers at this point, all with the aim of preserving the vehicle's attitude in a stable and nearly zero state throughout the entire flight duration.
There is an external fan in the vicinity of the hover point generating a wind disturbance with a speed of $4\;km/h$.
In the hover point, the vehicle resists the disturbance and holds its position.
Figure \ref{fig:error_position_ho} shows the position tracking error for each experimental trial. It is clear that, with all methods, the vehicle has successfully reached to hover point and held its position despite the disturbances. However, when compared to the other two methods, HGDO+SMC exhibits reduced fluctuations in the hover point.


{In this experiment, all the desired rotational states were set to zero. However, the external fan was generating disturbances in the rotational states. Fig. \ref{fig:Euler_angles_hold} showcases the attitude trajectory in each trial, confirming a better disturbance rejection for the rotational states using HGDO, especially with $\varepsilon = 0.01$. Table \ref{tab:hovering} compares the RMS of position and attitude tracking errors for different control strategies in this scenario.}



\begin{figure}[t]
    \centering
    \includegraphics[trim={0cm 0cm 0cm 0cm}, clip, width = 0.75\linewidth]{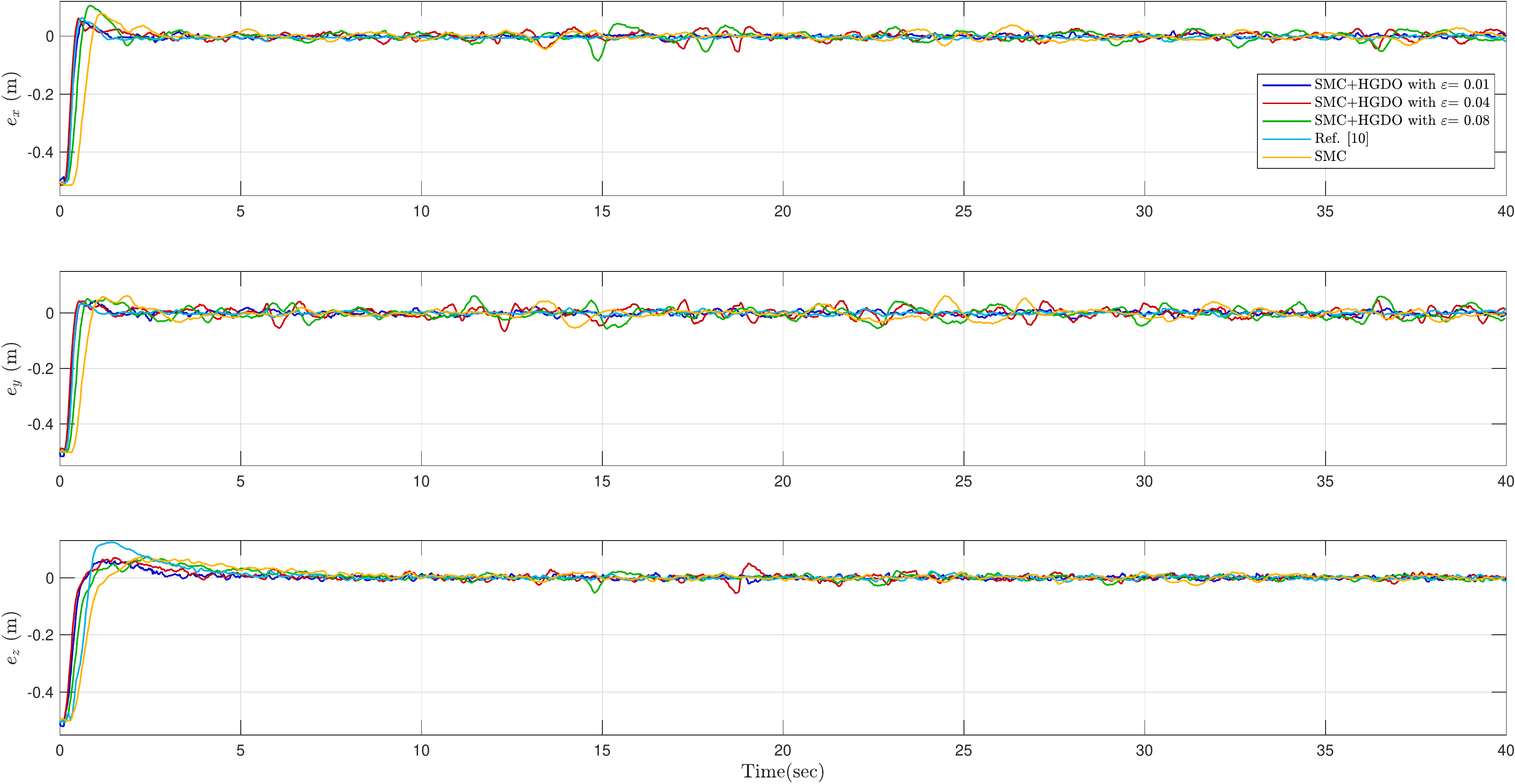}
    \caption{Position tracking error in the hovering scenario.}
    \label{fig:error_position_ho}
    \centering
    \includegraphics[trim={0cm 0cm 0cm 0cm}, clip, width = 0.75\linewidth]{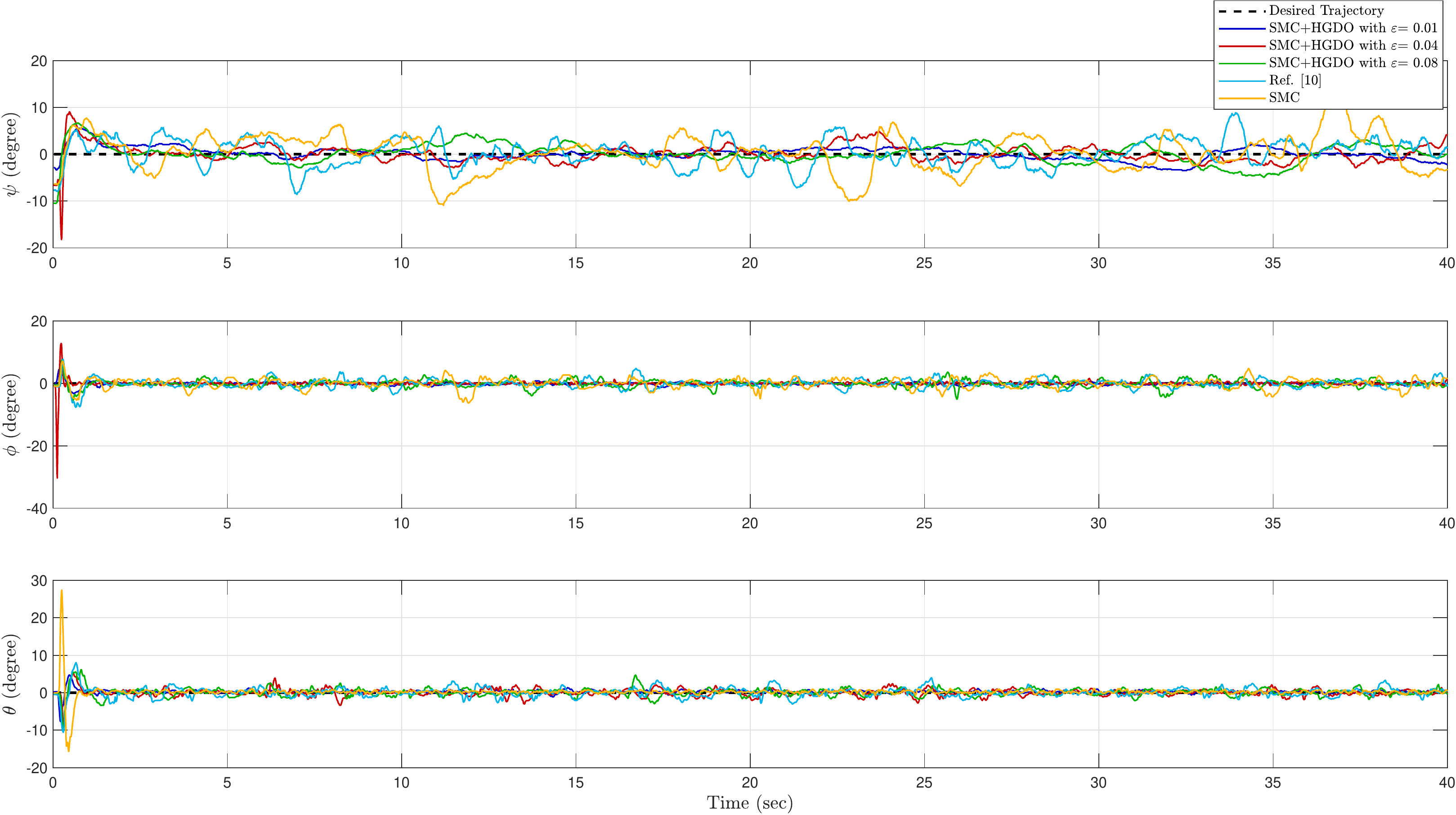}
    \caption{Desired and actual vehicle attitude in the hovering scenario.}
    \label{fig:Euler_angles_hold}
\end{figure}

\begin{table}[h]
  \caption{Root mean square of position and attitude tracking errors in the hovering scenario.}
  \label{tab:hovering}
  \resizebox{\textwidth}{!}{%
\begin{tabular}{@{}llllll@{}}
\toprule  
      Parameter & SMC+HGDO with $\varepsilon$= 0.01  & SMC+HGDO with $\varepsilon$= 0.04 & SMC+HGDO with $\varepsilon$= 0.08 & Ref. [10] & SMC \\
      \midrule
      $x$ & 0.042 & 0.044 & 0.055 & 0.044 & 0.065 \\
      $y$ & 0.043 & 0.044 & 0.055 &0.043 & 0.064 \\
      $z$ & 0.045 & 0.045 & 0.054 & 0.060 & 0.065 \\
      $\psi$ & 0.023 & 0.033 & 0.036 & 0.054 & 0.069 \\
      $\phi$ & 0.008 & 0.018 & 0.021 &0.024 & 0.027 \\
      $\theta$ & 0.010 & 0.016 & 0.019 & 0.022 & 0.026 \\ \bottomrule
      \end{tabular}
      }
\end{table}
\section{Conclusion}
This study developed an HGDO for robust trajectory tracking of quadrotors.
Our theoretical results established that (i) HGDO can guarantee the boundedness of disturbance estimation error with a short transient time, and (ii) HGDO combined with Lyapunov-based controllers can guarantee the boundedness of position and attitude tracking errors.
Our experimental results conformed with theoretical results and demonstrated that adding HGDO to a flight controller significantly improves the quadrotor's robustness against external disturbances. 

Note that despite HGDO's several desirable properties, its use in quadrotor control has remained minimal in the past.
Besides the fact that the advantages of HGDO were not thoroughly studied for quadrotors before this paper, there could be skepticism attributed to the sensitivity of conventional HGDOs to measurement noise.
However, our HGDO-based flight control was capable of handling the typical measurement noise present in inertial measurement units and motion capture systems in a common research vehicle. 
Therefore, we argue that HGDO is a simple, easy-to-tune, and computationally efficient module that can be added to conventional quadrotor flight control approaches to boost system robustness. 
The HGDO-based control performance can be further improved by explicitly considering measurement noise, high-frequency disturbances, and the limitations of sensors and actuators in the design. Future work in HGDO can delve into these topics to further strengthen HGDO advantages for quadrotor control.

\bibliographystyle{unsrtnat}
\bibliography{references}

\begin{thebibliography}{45}
\providecommand{\natexlab}[1]{#1}
\providecommand{\url}[1]{\texttt{#1}}
\expandafter\ifx\csname urlstyle\endcsname\relax
  \providecommand{\doi}[1]{doi: #1}\else
  \providecommand{\doi}{doi: \begingroup \urlstyle{rm}\Url}\fi

\bibitem[Galway et~al.(2008)Galway, Etele, and Fusina]{galway2008modeling}
David Galway, Jason Etele, and Giovanni Fusina.
\newblock Modeling of the urban gust environment with application to autonomous flight.
\newblock In \emph{AIAA Atmospheric Flight Mechanics Conference and Exhibit}, page 6565, 2008.
\newblock \doi{10.2514/6.2008-6565}.

\bibitem[He and Leang(2020)]{he2020quasi}
Xiang He and Kam~K Leang.
\newblock Quasi-steady in-ground-effect model for single and multirotor aerial vehicles.
\newblock \emph{AIAA Journal}, 58\penalty0 (12):\penalty0 5318--5331, 2020.
\newblock \doi{10.2514/1.J059223}.

\bibitem[Nathanael et~al.(2022)Nathanael, Wang, and Low]{nathanael2022numerical}
Joshua~C Nathanael, Chung-Hung~John Wang, and Kin~Huat Low.
\newblock Numerical studies on modeling the near-and far-field wake vortex of a quadrotor in forward flight.
\newblock \emph{Proceedings of the Institution of Mechanical Engineers, Part G: Journal of Aerospace Engineering}, 236\penalty0 (6):\penalty0 1166--1183, 2022.
\newblock \doi{10.1177/09544100211029074}.

\bibitem[Greeff and Schoellig(2018)]{greeff2018flatness}
Melissa Greeff and Angela~P Schoellig.
\newblock Flatness-based model predictive control for quadrotor trajectory tracking.
\newblock In \emph{2018 IEEE/RSJ International Conference on Intelligent Robots and Systems (IROS)}, pages 6740--6745. IEEE, 2018.
\newblock \doi{10.1109/IROS.2018.8594012}.

\bibitem[Zhang et~al.(2023)Zhang, Song, Song, and Stojanovic]{zhang2023finite}
Qiyuan Zhang, Xiaona Song, Shuai Song, and Vladimir Stojanovic.
\newblock Finite-time sliding mode control for singularly perturbed pde systems.
\newblock \emph{Journal of the Franklin Institute}, 360\penalty0 (2):\penalty0 841--861, 2023.
\newblock \doi{10.1016/j.jfranklin.2022.11.037}.

\bibitem[Nan et~al.(2022)Nan, Sun, Foehn, and Scaramuzza]{nan2022nonlinear}
Fang Nan, Sihao Sun, Philipp Foehn, and Davide Scaramuzza.
\newblock Nonlinear mpc for quadrotor fault-tolerant control.
\newblock \emph{IEEE Robotics and Automation Letters}, 7\penalty0 (2):\penalty0 5047--5054, 2022.
\newblock \doi{10.1109/LRA.2022.3154033}.

\bibitem[Xu and Ozguner(2006)]{xu2006sliding}
Rong Xu and Umit Ozguner.
\newblock Sliding mode control of a quadrotor helicopter.
\newblock In \emph{Proceedings of the 45th IEEE Conference on Decision and Control}, pages 4957--4962. IEEE, 2006.
\newblock \doi{10.1109/CDC.2006.377588}.

\bibitem[Hou et~al.(2020)Hou, Lu, and Tu]{hou2020nonsingular}
Zhiwei Hou, Peng Lu, and Zhangjie Tu.
\newblock Nonsingular terminal sliding mode control for a quadrotor uav with a total rotor failure.
\newblock \emph{Aerospace Science and Technology}, 98:\penalty0 105716, 2020.
\newblock \doi{10.1016/j.ast.2020.105716}.

\bibitem[Zheng et~al.(2014)Zheng, Xiong, and Luo]{zheng2014second}
En-Hui Zheng, Jing-Jing Xiong, and Ji-Liang Luo.
\newblock Second order sliding mode control for a quadrotor uav.
\newblock \emph{ISA transactions}, 53\penalty0 (4):\penalty0 1350--1356, 2014.
\newblock \doi{10.1016/j.isatra.2014.03.010}.

\bibitem[Kabiri et~al.(2019)Kabiri, Atrianfar, and Menhaj]{kabiri20193d}
Meisam Kabiri, Hajar Atrianfar, and Mohammad~Bagher Menhaj.
\newblock 3d trajectory tracking control for a thrust-propelled vehicle with time-varying disturbances.
\newblock \emph{International Journal of Control, Automation and Systems}, 17:\penalty0 1978--1986, 2019.
\newblock \doi{10.1007/s12555-018-0331-3}.

\bibitem[De~Monte and Lohmann(2013)]{de2013position}
Paul De~Monte and Boris Lohmann.
\newblock Position trajectory tracking of a quadrotor helicopter based on l1 adaptive control.
\newblock In \emph{2013 European Control Conference (ECC)}, pages 3346--3353. IEEE, 2013.
\newblock \doi{10.1515/auto-2013-1035}.

\bibitem[Song et~al.(2023)Song, Wu, Stojanovic, and Song]{song20231}
Xiaona Song, Chenglin Wu, Vladimir Stojanovic, and Shuai Song.
\newblock 1 bit encoding--decoding-based event-triggered fixed-time adaptive control for unmanned surface vehicle with guaranteed tracking performance.
\newblock \emph{Control Engineering Practice}, 135:\penalty0 105513, 2023.
\newblock \doi{10.1016/j.conengprac.2023.105513}.

\bibitem[Donti et~al.(2020)Donti, Roderick, Fazlyab, and Kolter]{donti2020enforcing}
Priya~L Donti, Melrose Roderick, Mahyar Fazlyab, and J~Zico Kolter.
\newblock Enforcing robust control guarantees within neural network policies.
\newblock \emph{arXiv preprint arXiv:2011.08105}, 2020.
\newblock \doi{10.48550/arXiv.2011.08105}.

\bibitem[Han et~al.(2022)Han, Cheng, Xi, and Yao]{han2022cascade}
Haoran Han, Jian Cheng, Zhilong Xi, and Bingcai Yao.
\newblock Cascade flight control of quadrotors based on deep reinforcement learning.
\newblock \emph{IEEE Robotics and Automation Letters}, 7\penalty0 (4):\penalty0 11134--11141, 2022.
\newblock \doi{10.1109/LRA.2022.3196455}.

\bibitem[Youcef-Toumi and Ito(1988)]{youcef1990time}
Kamal Youcef-Toumi and Osamu Ito.
\newblock A time delay controller for systems with unknown dynamics.
\newblock In \emph{1988 American Control Conference}, pages 904--913, 1988.
\newblock \doi{10.23919/ACC.1988.4789852}.

\bibitem[Lim and Jung(2014)]{lim2014altitude}
Jeong~Geun Lim and Seul Jung.
\newblock Altitude control of a quad-rotor system by using a time-delayed control method.
\newblock \emph{Journal of Institute of Control, Robotics and Systems}, 20\penalty0 (7):\penalty0 724--729, 2014.
\newblock \doi{10.5302/J.ICROS.2014.13.1947}.

\bibitem[Hall and Shtessel(2006)]{hall2006sliding}
Charles~E Hall and Yuri~B Shtessel.
\newblock Sliding mode disturbance observer-based control for a reusable launch vehicle.
\newblock \emph{Journal of guidance, control, and dynamics}, 29\penalty0 (6):\penalty0 1315--1328, 2006.
\newblock \doi{10.2514/1.20151}.

\bibitem[Shi et~al.(2018)Shi, Wu, and Chou]{shi2018generalized}
Di~Shi, Zhong Wu, and Wusheng Chou.
\newblock Generalized extended state observer based high precision attitude control of quadrotor vehicles subject to wind disturbance.
\newblock \emph{IEEE Access}, 6:\penalty0 32349--32359, 2018.
\newblock \doi{10.1109/ACCESS.2018.2842198}.

\bibitem[Talole and Phadke(2008)]{talole2008model}
SE~Talole and SB~Phadke.
\newblock Model following sliding mode control based on uncertainty and disturbance estimator.
\newblock \emph{ASME Journal of Dynamic Systems, Measurement, and Control}, 130\penalty0 (3):\penalty0 034501, 2008.

\bibitem[Chandar and Talole(2014)]{chandar2014improving}
TS~Chandar and SE~Talole.
\newblock Improving the performance of ude-based controller using a new filter design.
\newblock \emph{Nonlinear Dynamics}, 77\penalty0 (3):\penalty0 753--768, 2014.
\newblock \doi{10.1007/s11071-014-1337-x}.

\bibitem[Zhong and Rees(2004)]{zhong2004control}
Qing-Chang Zhong and David Rees.
\newblock Control of uncertain lti systems based on an uncertainty and disturbance estimator.
\newblock \emph{J. Dyn. Sys., Meas., Control}, 126\penalty0 (4):\penalty0 905--910, 2004.
\newblock \doi{10.1115/1.1850529}.

\bibitem[Zhou et~al.(2022)Zhou, Tao, Chen, Stojanovic, and Paszke]{zhou2022robust}
Chenhui Zhou, Hongfeng Tao, Yiyang Chen, Vladimir Stojanovic, and Wojciech Paszke.
\newblock Robust point-to-point iterative learning control for constrained systems: A minimum energy approach.
\newblock \emph{International Journal of Robust and Nonlinear Control}, 32\penalty0 (18):\penalty0 10139--10161, 2022.
\newblock \doi{10.1002/rnc.6354}.

\bibitem[Liang et~al.(2021)Liang, Chen, and Yao]{liang2021geometric}
Weisheng Liang, Zheng Chen, and Bin Yao.
\newblock Geometric adaptive robust hierarchical control for quadrotors with aerodynamic damping and complete inertia compensation.
\newblock \emph{IEEE Transactions on Industrial Electronics}, 69\penalty0 (12):\penalty0 13213--13224, 2021.
\newblock \doi{10.1109/TIE.2021.3137615}.

\bibitem[Fethalla et~al.(2018)Fethalla, Saad, Michalska, and Ghommam]{fethalla2018robust}
Nuradeen Fethalla, Maarouf Saad, Hannah Michalska, and Jawhar Ghommam.
\newblock Robust observer-based dynamic sliding mode controller for a quadrotor uav.
\newblock \emph{IEEE access}, 6:\penalty0 45846--45859, 2018.
\newblock \doi{10.1109/ACCESS.2018.2866208}.

\bibitem[Huang et~al.(2022)Huang, Huang, Qin, Li, and Yang]{huang2022finite}
Deqing Huang, Tianpeng Huang, Na~Qin, Yanan Li, and Yong Yang.
\newblock Finite-time control for a uav system based on finite-time disturbance observer.
\newblock \emph{Aerospace Science and Technology}, 129:\penalty0 107825, 2022.
\newblock \doi{10.1016/j.ast.2022.107825}.

\bibitem[Stojanovi{\'c}(2023)]{stojanovic2023fault}
Vladimir Stojanovi{\'c}.
\newblock Fault-tolerant control of a hydraulic servo actuator via adaptive dynamic programming.
\newblock \emph{Mathematical Modelling and Control}, 2023.
\newblock \doi{10.3934/mmc.2023016}.

\bibitem[Khalil(2017{\natexlab{a}})]{khalil2017high}
Hassan~K Khalil.
\newblock \emph{High-gain observers in nonlinear feedback control}.
\newblock SIAM, 2017{\natexlab{a}}.
\newblock \doi{10.1002/rnc.3051}.

\bibitem[Boizot et~al.(2010)Boizot, Busvelle, and Gauthier]{boizot2010adaptive}
Nicolas Boizot, Eric Busvelle, and Jean-Paul Gauthier.
\newblock An adaptive high-gain observer for nonlinear systems.
\newblock \emph{Automatica}, 46\penalty0 (9):\penalty0 1483--1488, 2010.
\newblock \doi{10.1016/j.automatica.2010.06.004}.

\bibitem[Won et~al.(2015)Won, Kim, Shin, and Chung]{won2015high}
Daehee Won, Wonhee Kim, Donghoon Shin, and Chung~Choo Chung.
\newblock High-gain disturbance observer-based backstepping control with output tracking error constraint for electro-hydraulic systems.
\newblock \emph{IEEE Transactions on Control Systems Technology}, 23\penalty0 (2):\penalty0 787--795, 2015.
\newblock \doi{10.1109/TCST.2014.2325895}.

\bibitem[Khalil(2017{\natexlab{b}})]{khalil2017extended}
Hassan~K Khalil.
\newblock Extended high-gain observers as disturbance estimators.
\newblock \emph{SICE Journal of Control, Measurement, and System Integration}, 10\penalty0 (3):\penalty0 125--134, 2017{\natexlab{b}}.
\newblock \doi{10.9746/jcmsi.10.125}.

\bibitem[Boss et~al.(2017)Boss, Lee, and Choi]{boss2017uncertainty}
Connor~J Boss, Joonho Lee, and Jongeun Choi.
\newblock Uncertainty and disturbance estimation for quadrotor control using extended high-gain observers: Experimental implementation.
\newblock In \emph{Dynamic Systems and Control Conference}, volume 58288, page V002T01A003. American Society of Mechanical Engineers, 2017.
\newblock \doi{10.1115/DSCC2017-5204}.

\bibitem[Zhao et~al.(2019)Zhao, Zhang, Ma, and Xia]{zhao2019composite}
Kai Zhao, Jinhui Zhang, Dailiang Ma, and Yuanqing Xia.
\newblock Composite disturbance rejection attitude control for quadrotor with unknown disturbance.
\newblock \emph{IEEE Transactions on Industrial Electronics}, 67\penalty0 (8):\penalty0 6894--6903, 2019.
\newblock \doi{10.1109/TIE.2019.2937065}.

\bibitem[Lu et~al.(2020)Lu, Ren, and Parameswaran]{lu2020uncertainty}
Qi~Lu, Beibei Ren, and Siva Parameswaran.
\newblock Uncertainty and disturbance estimator-based global trajectory tracking control for a quadrotor.
\newblock \emph{IEEE/ASME Transactions on Mechatronics}, 25\penalty0 (3):\penalty0 1519--1530, 2020.
\newblock \doi{10.1109/TMECH.2020.2978529}.

\bibitem[Boss and Srivastava(2021)]{boss2021high}
Connor~J Boss and Vaibhav Srivastava.
\newblock A high-gain observer approach to robust trajectory estimation and tracking for a multi-rotor uav.
\newblock \emph{arXiv preprint arXiv:2103.13429}, 2021.
\newblock \doi{10.48550/arXiv.2103.13429}.

\bibitem[Lee et~al.(2021)Lee, Seo, and Choi]{lee2021output}
Joonho Lee, Joohwan Seo, and Jongeun Choi.
\newblock Output feedback control design using extended high-gain observers and dynamic inversion with projection for a small scaled helicopter.
\newblock \emph{Automatica}, 133:\penalty0 109883, 2021.
\newblock \doi{10.1016/j.automatica.2021.109883}.

\bibitem[Fang et~al.(2016)Fang, Wu, Shang, and Dong]{fang2016robust}
Xing Fang, Aiguo Wu, Yujia Shang, and Na~Dong.
\newblock Robust control of small-scale unmanned helicopter with matched and mismatched disturbances.
\newblock \emph{Journal of the Franklin Institute}, 353\penalty0 (18):\penalty0 4803--4820, 2016.
\newblock \doi{10.1016/j.jfranklin.2016.09.016}.

\bibitem[Mahony et~al.(2012)Mahony, Kumar, and Corke]{mahony2012multirotor}
Robert Mahony, Vijay Kumar, and Peter Corke.
\newblock Multirotor aerial vehicles: Modeling, estimation, and control of quadrotor.
\newblock \emph{IEEE robotics \& automation magazine}, 19\penalty0 (3):\penalty0 20--32, 2012.
\newblock \doi{10.1109/MRA.2012.2206474}.

\bibitem[Bouabdallah and Siegwart(2007)]{bouabdallah2007full}
Samir Bouabdallah and Roland Siegwart.
\newblock Full control of a quadrotor.
\newblock In \emph{2007 IEEE/RSJ international conference on intelligent robots and systems}, pages 153--158. Ieee, 2007.
\newblock \doi{10.1109/IROS.2007.4399042}.

\bibitem[Bouabdallah and Siegwart(2006)]{bouabdallah2006towards}
Samir Bouabdallah and Roland Siegwart.
\newblock Towards intelligent miniature flying robots.
\newblock In \emph{Field and Service Robotics: Results of the 5th International Conference}, pages 429--440. Springer, 2006.
\newblock \doi{10.1007/978-3-540-33453-8_36}.

\bibitem[Martini et~al.(2022)Martini, S{\"o}nmez, Rizzo, Stefanovic, Rutherford, and Valavanis]{martini2022euler}
Simone Martini, Serhat S{\"o}nmez, Alessandro Rizzo, Margareta Stefanovic, Matt~J Rutherford, and Kimon~P Valavanis.
\newblock Euler-lagrange modeling and control of quadrotor uav with aerodynamic compensation.
\newblock In \emph{2022 International Conference on Unmanned Aircraft Systems (ICUAS)}, pages 369--377. IEEE, 2022.
\newblock \doi{10.1109/ICUAS54217.2022.9836215}.

\bibitem[Wheeden(2015)]{wheeden2015measure}
Richard~L Wheeden.
\newblock \emph{Measure and integral: an introduction to real analysis}, volume 308.
\newblock CRC press, 2015.

\bibitem[cra(2023)]{crazyflie}
Crazyflie 2.1, 2023.
\newblock URL \url{https://store.bitcraze.io/products/crazyflie-2-1}.

\bibitem[Beal(1993)]{beal1993digital}
TR~Beal.
\newblock Digital simulation of atmospheric turbulence for dryden and von karman models.
\newblock \emph{Journal of Guidance, Control, and Dynamics}, 16\penalty0 (1):\penalty0 132--138, 1993.
\newblock \doi{10.2514/3.11437}.

\bibitem[Ahrens and Khalil(2009)]{ahrens2009high}
Jeffrey~H Ahrens and Hassan~K Khalil.
\newblock High-gain observers in the presence of measurement noise: A switched-gain approach.
\newblock \emph{Automatica}, 45\penalty0 (4):\penalty0 936--943, 2009.
\newblock \doi{10.1016/j.automatica.2008.11.012}.

\bibitem[Sanfelice and Praly(2011)]{sanfelice2011performance}
Ricardo~G Sanfelice and Laurent Praly.
\newblock On the performance of high-gain observers with gain adaptation under measurement noise.
\newblock \emph{Automatica}, 47\penalty0 (10):\penalty0 2165--2176, 2011.
\newblock \doi{10.1016/j.automatica.2011.08.002}.

\end{thebibliography}

\end{document}